\documentclass{article}

\pdfoutput=1

\usepackage{arxiv}
\usepackage{amsfonts}
\usepackage{xcolor}
\usepackage{times}
\usepackage{soul}
\usepackage{url}
\usepackage{caption}
\usepackage{graphicx}
\usepackage{amsmath}
\usepackage{amsthm}
\usepackage{booktabs}
\usepackage{mathtools}
\usepackage{subcaption}
\usepackage{varwidth}
\usepackage{wrapfig}
\usepackage{include/eegl_macros}
\usepackage{xcolor}
\usepackage[inline]{enumitem}
\usepackage[export]{adjustbox}
\usepackage[numbers]{natbib}
\usepackage{doi}

\usepackage{algorithm}
\usepackage[noend]{algpseudocode}
\usepackage{ulem}
\algnewcommand\algorithmicinput{\textbf{Input:}}
\algnewcommand\Input{\item[\algorithmicinput]}
\algnewcommand\algorithmicoutput{\textbf{Output:}}
\algnewcommand\Output{\item[\algorithmicoutput]}
\algnewcommand\algorithmicparameters{\textbf{Parameters:}}
\algnewcommand\Parameters{\item[\algorithmicparameters]}

\newcommand{\cD}{\mathcal{D}}

\newcommand{\Gaston}{{\sc Gaston}}
\newcommand{\GNNExplainer}{{\sc GNNExplainer}}
\newcommand{\EEGL}{EEGL}
\newcommand{\setofclasslabels}{C}
\newcommand{\classlabel}{c}

\newcommand{\featurevector}{\vec{x}}
\newcommand{\featurematrix}{X}

\newcommand{\CA}{\textrm{C}_{60}}
\newcommand{\CB}{\textrm{C}_{70}}
\newcommand{\GM}[1]{G(M_#1)}
\newcommand{\GMp}[1]{G(M_{#1}')}
\newcommand{\shadowgnn}{{\sc shaDow-GNN}}

\title{Iterative Graph Neural Network Enhancement via Frequent Subgraph Mining of Explanations}
\author{
\href{https://orcid.org/0000-0002-1420-2688}{\includegraphics[scale=0.06]{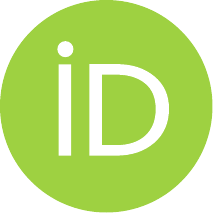}\hspace{1mm}Harish G. Naik} \\
Department of Computer Science\\
University of Illinois at Chicago\\
hnaik2@uic.edu \thanks{This publication is based on work supported in part by the University of Illinois at Chicago and the National Science Foundation (NSF) award \#CNS-1828265 for MRI: Acquisition of a Composable Platform as a Service Instrument for Deep Learning \& Visualization (COMPaaS DLV).}
\And
Jan Polster\\
Department of Computer Science \\
University of Bonn\\
s6japols@uni-bonn.de\\
\And
Raj Shekhar\\
Department of Computer Science\\
University of Illinois at Chicago\\
rshekh3@uic.edu \thanks{Supported by NSF grants 1828265 and 2217023}\\
\And
Tamás Horváth\\
Dept. of Computer Science, University of Bonn\\
Fraunhofer IAIS, Sankt Augustin\\ 
Lamarr Institute for \\
Machine Learning and Artificial Intelligence \\
horvath@cs.uni-bonn.de \thanks{This research has been funded by the Federal Ministry of Education and Research of Germany and the state of North Rhine-Westphalia as part of the Lamarr Institute for Machine Learning and Artificial Intelligence.}\\
\And
György Turán\\
University of Illinois at Chicago\\
HUN-REN-SZTE Research Group on AI, Szeged\\
gyt@uic.edu \thanks{Partially supported by NSF grants 2217023, 2240532 and by the AI National Laboratory Program (RRF-2.3.1-21-2022-00004).}
}

\begin{document}

\newcommand{\fix}{\marginpar{FIX}}
\newcommand{\new}{\marginpar{NEW}}

\maketitle


\begin{abstract}

We formulate an XAI-based model improvement approach
for Graph Neural Networks (GNNs) for node classification, called Explanation Enhanced Graph Learning (EEGL). The goal is to improve 
predictive performance of GNN using explanations. 
EEGL is an iterative self-improving algorithm, which
starts with a learned ``vanilla'' GNN, 
and repeatedly uses 
frequent subgraph mining to find relevant patterns in explanation subgraphs. These patterns are then filtered further to obtain application-dependent features corresponding to the presence of certain subgraphs in the node neighborhoods. Giving an application-dependent algorithm for such a subgraph-based extension of the Weisfeiler-Leman (1-WL) algorithm has previously been posed as an open problem. 
We present experimental evidence, with synthetic and real-world data, which show that {\EEGL} outperforms related approaches in predictive performance and that it has a node-distinguishing power beyond that of vanilla GNNs. We also analyze {\EEGL}'s training dynamics.
\end{abstract}

\section{Introduction}
\label{sec:introduction}

XAI-based model improvement is a relatively recent research direction in XAI. The underlying observation is that explanations, besides providing information to the user about the model's decision, can also be used to improve the model.
A recent survey is given in~\cite{Weber23}. 

In this paper we consider Graph Neural Networks (GNNs) for node classification within the XAI-based model improvement paradigm. 
Message-passing GNNs (MPNNs)\footnote{
A note on graph neural network terminology:
GNN is the general term,~\citep{barcelo21-graph_params} uses MPNN and~\citep{kipf17-gcn}
uses GCN. We use these terms depending on context.} have limited representational power due to the limitations of the related Weisfeiler-Leman (1-WL) algorithm~\citep{Morris19,xu19-gin}. Two nodes which are indistinguishable by 1-WL are indistinguishable by any MPNN. 

Extending the power of MPNN to overcome these limitations is a fundamental challenge.
One approach is higher-order variants of WL, which are more expressive, but computationally less tractable~\cite{Dwivedi23}.
Another approach is to add \emph{structural information as node features}. For vanilla MPNN node feature vectors are constant. The number of triangles in the 1-hop neighborhood of a node, which cannot be detected by 1-WL, can be such an added feature. 
In~\citep{Bour23}, the authors propose adding node features by counting the number of rooted subgraphs that are isomorphic to a given rooted pattern graph.
They suggest domain-specific patterns, e.g., cliques for social networks and cycles for molecules. Adding new features by counting the number of rooted homomorphisms 
is proposed in~\citep{barcelo21-graph_params}.
Building on these proposals, a general question,
formulated in~\citep{barcelo21-graph_params}, is
how to select useful patterns? We refer to this problem as the \emph{GNN pattern selection problem}.
It is noted in~\citep{barcelo21-graph_params}
that ``one important remark is that selecting the best set of features is still a challenging endeavor''. They also point out that the \emph{application-dependence} of the best set of features adds to the difficulty of the problem.

In this paper we propose an approach to the GNN pattern selection problem for \textit{application-dependent} features, i.e., the features are extracted from the data in an automatic way, and \textit{not} by using some domain-specific pattern set.
XAI-based model improvement suggests to use explanations for solving the GNN pattern selection problem.
Utilizing that explanations for GNNs are often \emph{rooted subgraphs}, we propose the {\sc Explanation Enhanced Graph Learning} (EEGL) approach for iterative enhancement of the predictive performance of GNNs for node classification. 
{\EEGL} uses \emph{frequent connected subgraph mining} to identify patterns occuring in many explanation subgraphs for nodes of a given class. These are analyzed further to select the best patterns. Nodes are then annotated with features based on the presence or absence of these patterns in the neighborhood. Starting with vanilla GNN, the process can be iterated. 

A few remarks are in order about assumptions underlying the approach and its intended scope of applications.
It is assumed that the learning problem is such that 1. \emph{there exist explanation subgraphs of reasonable quality} and 2. \emph{the GNN learning and explanation algorithm to be improved is of reasonable quality}. 
The \textit{hypothesis} underlying our work, then, is that if assumptions 1. and 2. hold then explanations can be used to construct useful features, and those can be used to improve performance. Assumptions 1. and 2. are \emph{not} expected to hold in general. On the other hand, assumption 2. is not too restrictive as the goal is \textit{to improve a weak learner using weak explanations}, a setup which may be of interest in itself.
Applications where the assumptions may hold could be in \emph{scientific discovery}.
A scientist could suggest a kind of explanation to the
data scientist, who could use EEGL to evaluate, use or correct this suggestion~\citep{naik21-expspec}.

For the experimental evaluation of the EEGL approach, one can consider the following questions:

\begin{description}
    \item Q1:\label{q:predictiveperformance} Can the node-distinguishing power and hence, the predictive performance of GNNs be improved by using frequent subgraph mining of explanations? Do iterations help?
    \item Q2: \label{q:featuredefinitions} How does the performance of 
    {\EEGL}-trained GNNs compare to other 
    feature initializations and subgraph-based GNNs?
    \item Q3: \label{q:symmetries} What is the effect of the relationship between 
    1-WL symmetries and label partitions?
\end{description}

\textbf{Summary of the Experimental Results}
The objective of the paper is to provide first steps towards answering these questions, focusing
on the effect of \textit{structural complexity}. One should note that our approach is \emph{not} expected to be applicable in cases when the information provided by the graph is more of a statistical nature, as in social networks.
In contrast, motivated by the connections to 1-WL,
we consider synthetic and real-world datasets with different relationships between the node classes having the same 1-WL labels and those having the same class labels. 
From this point of view, datasets studied belong
to three categories:
(i) The graph has a large number of 1-WL classes, compared to its size (\textit{low} 1-WL symmetry) and target classes are formed by the union of a family of 1-WL classes, 
(ii) the graph has a small number of 1-WL classes, compared to its size (\textit{high} 1-WL symmetry) and all 1-WL classes are formed by the union of a family of target classes, and its special case that 
(iii) the graph is regular (each node has the same degree) and has therefore only one 1-WL class.
We used synthetic datasets for (i) and (ii), and \textit{fullerenes} for (iii).
The following are some of the main conclusions of our experiments.
\begin{description}
\item Answer to Q1: 
{\EEGL} was capable of iterative self-improvement from explanations in at most 3 iterations on \textit{all} learning tasks considered, including those that require a node-distinguishing power beyond that of vanilla GNNs.
\item Answer to Q2: We considered three other feature initializations:
(1) one-hot encoding of true class labels (regarded as an upper bound on the performance of GNN on the learning task),
(2) randomly assigned numerical features ~\cite{Abbou21,Dasou}, and
(3) ``maliciously'' selected subgraph features (motivated by the
comment of~\citep{barcelo21-graph_params} that performance ``almost always benefits from any set of additional features''). 
We also compared {\EEGL}'s predictive performance to that of \shadowgnn~\cite{Zeng_etal/2021}, 
a subgraph-based approach which also generates node representations
\textit{dynamically}.
{\EEGL} \textit{outperforms} (2) and (3) and {\shadowgnn} on all datasets, except for one, where {\shadowgnn} achieved approximately the same F1-score. 
\item Answer to Q3: {\EEGL} performed well on datasets of all three types.
Its predictive performance was close to 100\% on all problems in (i) and (ii). 
Regarding (iii), {\EEGL} achieved 100\%  on most and between 86 and 97\% on 3 fullerenes, most in the first iteration. 
It was, however, unable to improve the results on the 3 fullerenes in the subsequent rounds.  
{\EEGL} achieved already  more than 99\% in average in one iteration on all problems in category (i) and 97\% in category (iii). For the dataset in category (ii), it needed three iterations to achieve 100\%. 
\end{description}

\textbf{Contributions} 
In summary, the main contribution of this work is a general \textit{self-improving} framework to overcome the inherent limited representational power of vanilla GNNs. In particular, it 
(i) applies an XAI-based iterative model improvement method to GNN, (ii) provides an automated, application-dependent solution to the GNN pattern selection problem, (iii) applies frequent connected subgraph mining to XAI, and (iv) gives a detailed study of node classification problems where graph-theoretic structural complexity is relevant.

\textbf{Outline} The rest of the paper is organized as follows. 
In Sections \ref{sec:related} we overview related work.
The {\EEGL} system is presented in Section~\ref{sec:EEGL}.
We report and discuss the experimental results obtained in Section~\ref{sec:experiments}. 
Finally, in Section~\ref{sec:conclusion} we conclude and formulate some problems for further research.

\section{Related Work}
\label{sec:related}

Besides the most closely related works, already cited in Section~\ref{sec:introduction}, we briefly mention some further references, using survey papers covering related work when possible. A recent general survey of XAI, including some discussion of the role of XAI-based model improvement, is \cite{Ali23}. In the survey of
\cite{MorrisGo23} of the different variations of the Weisfeiler-Leman algorithm, Section 5.2 is about neural architectures extending 1-WL, including~\citep{barcelo21-graph_params,Bour23}.
A subsection discusses other subgraph-enhanced approaches, including {\shadowgnn}~\cite{Zeng_etal/2021} on node classification, which implement symmetry-breaking to address the limitations of 1-WL (see Appendix~\ref{sec:shadowgnn}). Thus these approaches exploit properties of the network, but do \textit{not} take the learning problem into consideration (with the exception of \cite{Zeng_etal/2021}). A detailed computational study is given in~\cite{Dwivedi23}, including results on $k$-WL. The need for synthetic examples is discussed in
~\cite{Maekawa22}.


Explanatory interactive machine learning \citep{Fried22,Fried23} considers explanations returned by the learner and corrected by the user as part of the human-in-the-loop learning process. Faithfulness, a basic metric for evaluating explanations, measures the predictive power of explanation subgraphs~\citep{AgarZ22}~\footnote{Another comment on terminology: in this paper we do not discuss the evaluation of explanations. Thus we mean accuracy in the standard ML, and not in the XAI sense, as used in \cite{AgarZ22} and somewhat differently in~\cite{ying19-gnn_explainer}.}.
Other works, closer to the present paper, use explanations in an automated manner as a tool for improving prediction. 
In \cite{Shedi3} the effect of incorporating linear approximations into the learning process is considered. Similarly, \cite{AgarQ23} uses explainability information to guide message passing in GNN. The SUGAR system~\cite{Sun021} uses a reinforcement pooling mechanism to incorporate significant subgraphs into graph classification.

\section{The EEGL System}
\label{sec:EEGL}


Before presenting the main components of the {\sc Explanation Enhanced Graph Learning} (\EEGL) system, we first recall some necessary notions. 
Background on GNN, {\GNNExplainer}, Weisfeiler-Leman algorithm, and frequent subgraph mining is given in \cite{hamilton17-representation,nijssen05-gaston,ying19-gnn_explainer} and Appendix~\ref{sec:background}. 
%

\textbf{Notions and Notation}
For a graph $G$, $V(G)$ and $E(G)$ denote the sets of nodes and edges. Graphs are always undirected.
A \emph{rooted graph}
is a pair $(G, v)$, where $v \in V(G)$.
Graph $G_1$ is \textit{isomorphic} to graph $G_2$ if there is a \textit{bijection} $\psi: V(G_1) \to V(G_2)$ such that
$\{u,v\}\in E(G_1)$ iff $\{\psi(u),\psi(v)\} \in E(G_2)$ for all $u,v\in V(G_1)$. 
Furthermore, $G_1$ is \textit{subgraph} \textit{isomorphic} to $G_2$ if $G_2$ has a  subgraph that is isomorphic to $G_1$, or equivalently, if there exists an \textit{injective} function $\varphi: V(G_1) \to V(G_2)$ 
such that $\{\varphi(u),\varphi(v)\} \in E(G_2)$ if $\{u,v\}\in E(G_1)$, for all $u,v\in V(G_1)$.
A rooted graph $(G_1, r)$ has a \textit{rooted subgraph isomorphism} 
into a rooted graph $(G_2, v)$ if there is a subgraph isomorphism $\varphi$ from $G_1$ into $G_2$ such that $\varphi(r) = v$.
A \textit{rooted pattern} is a pair $(P,v)$ such that $P$ is a connected graph and $v \in V(P)$.
The 1-WL algorithm~\cite{hamilton17-representation} defines a partition of $E(G)$ of a graph $G$. The blocks of this partition will be referred to as the \textit{1-WL classes} of $G$. 
Nodes $u,v$ of $G$ belong to the same \textit{orbit} of $G$ iff there is an automorphism (i.e., isomorphism from $G$ to $G$) mapping $u$ to $v$. The orbits define an equivalence relation over $E(G)$, which is a refinement of the equivalence relation defined by the 1-WL classes.

\begin{algorithm}[t]
	\begin{algorithmic}[1]
        \Parameters feature vector dimension $d$, iteration number $K$, set $C$ of class labels
		\Input graph $G$ with $n$ vertices, set $T \subseteq V(G) \times C$ of training examples, relative frequency threshold $\tau \in (0,1]$
		\Output GNN model $\Phi: V(G) \to C$ 
		\Statex 
		\State $X \gets \text{\sc Init\_Feature\_Matrix}(G,d)$
		\label{line:init_feature_matrix}
		\For{$k=1,\ldots,K$} \label{line:start_for_loop}
		\State $\Phi \gets \text{\sc GNN\_Learning}(G,X,T)$ 
        \label{line:GNN_learning_module}
		\State \textbf{for all} $c \in C$ \textbf{do} $\mathcal{E}_c \gets \emptyset$ 
		\label{line:start_node_explainer_module}
		\State \textbf{for all} $v \in V(G)$ \label{line:node_explainer_loop}
		\State \quad $E_v \gets \text{\sc GNN\_Node\_Explainer}(G,X,\Phi,v)$
		\label{line:node_explanation}
		\State \quad add $E_v$ to $\mathcal{E}_c$, where $c = \Phi(v)$ 
		\label{line:end_node_explainer_module}
		\ForAll{$c \in C$} 
		\label{line:start_pattern_extraction}
		\State $\mathcal{P}_c \gets \text{\sc Maximal\_Frequent\_Patterns}(\mathcal{E}_c, \tau)$
		\label{line:maximal_frequent_patterns}
		\EndFor
		\State $\mathcal{P}^\top \gets \text{\sc Top\_Patterns}(\mathcal{P}, \min \{d, |\mathcal{P}|\}, T)$, where \newline \hspace*{6em} $\mathcal{P} =\bigcup_{c\in C} \mathcal{P}_c$ 
		\label{line:select_top_patterns}
		\State $X \gets \text{\sc Update\_Feature\_Matrix}(G,\Phi,\mathcal{P}^\top, d)$
        \label{line:feature_annotation}
		\EndFor 
		\State \Return $\Phi$
	\end{algorithmic}
	\caption{\sc Explanation Enhanced Graph Learning} 
	\label{alg:EEGL}
\end{algorithm}

\textbf{The {\EEGL} System}
The pseudocode of {\EEGL} is given in 
Alg.~\ref{alg:EEGL} (see, also, Fig.~\ref{fig:eegl} in Appendix~\ref{sec:eegl_app} for a high-level depiction)
It consists of the 
(i) GNN learning, 
(ii) node explainer, 
(iii) pattern extraction, and
(iv) feature annotation
modules.
The four modules 
are used to learn an unknown target function $f: V(G) \to \setofclasslabels$, where 
$G$ is the input graph and 
$\setofclasslabels$ is a finite set of class labels.
In addition to $G$, the input to {\EEGL} also contains a set $T = \{ (v, f(v)): v \in V'\}$ of training examples for some $V' \subseteq V(G)$, the dimension $d$ of the node feature vectors, a relative frequency threshold $\tau \in (0,1]$, and a positive integer $K$ specifying the number of iterations (cf. Alg.~\ref{alg:EEGL}). 
In each iteration of the outer loop in Alg.~\ref{alg:EEGL} (lines~\ref{line:start_for_loop}--\ref{line:feature_annotation}), all nodes of $G$ are associated with a $d$-dimensional \textit{feature vector}.
The feature vectors of the nodes are represented by an $n \times d$ \textit{feature matrix} $\featurematrix$, where $n = |V(G)|$.
We now describe the above four modules.


\textbf{GNN Learning Module (line~\ref{line:GNN_learning_module} of Alg.~\ref{alg:EEGL})}
In each iteration, {\EEGL} first learns a new GNN 
$\Phi$ for $G$ using $G$, 
$\featurematrix$, and $T$ as input.
While $G$ and $T$ are unchanged, 
$\featurematrix$ is recalculated in each iteration. 
It is initialized in line~\ref{line:init_feature_matrix} and updated in line~\ref{line:feature_annotation}.
In case of the ``vanilla'' initialization, $\featurematrix$ is set to the $n \times d$ matrix of ones. 
The function {\sc GNN\_Learning}  (line~\ref{line:GNN_learning_module}) is realized with 
GCN~\citep{kipf17-gcn}.

\textbf{Node Explainer Module (lines~\ref{line:start_node_explainer_module}--\ref{line:end_node_explainer_module} of Alg.~\ref{alg:EEGL}})  
The GNN model $\Phi$ is used as input to the \textit{node explainer} module, together with $G$ and $\featurematrix$. 
It is called for each $v \in V(G)$ separately (lines~\ref{line:node_explainer_loop}--\ref{line:node_explanation}) and  
returns an \textit{individual} subgraph $E_v$ of $G$ as the explanation for the model's prediction of the class of $v$ by $\Phi(v)$.
%
In our experiments, the node explainer function {\sc GNN\_Node\_Explainer} (line~\ref{line:node_explanation}) is realized with the {\GNNExplainer} system~\citep{ying19-gnn_explainer}.
The 
explanation graphs returned by {\GNNExplainer} contained the nodes themselves.\footnote{Parameters can be chosen so that \GNNExplainer~\cite{ying19-gnn_explainer} is forced to include $v$ in $E_v$ for \textit{all} $v \in V(G)$.}
Node $v$ is marked as the \textit{root} of $E_v$.
We note that {\GNNExplainer} also calculates a feature mask for each explanation pattern.
These are disregarded in EEGL but could play a role in future work. 
The explanation graphs are partitioned according to the  class labels \textit{predicted} by $\Phi$ (lines~\ref{line:start_node_explainer_module} and \ref{line:end_node_explainer_module}); the block containing the explanation graphs for a class $c \in C$ is denoted by $\mathcal{E}_c$ (cf. line~\ref{line:end_node_explainer_module}).

\textbf{Pattern Extraction Module (lines~\ref{line:start_pattern_extraction}--\ref{line:select_top_patterns})}
This module generates a set of 
\textit{maximal frequent} \textit{rooted} patterns from the explanation graphs computed by the previous module. 
This 
will then be used by the next module.
The underlying assumption behind the synthetic and real-world examples 
is that there is a set of \textit{class patterns} for each class label.
More precisely, for input graph $G$, target function $f$, and class label $\classlabel \in \setofclasslabels$ 
there is a set $S_\classlabel$ of (almost) \textit{contrastive} rooted patterns such that (i) for most~\footnote{The qualifications \emph{almost, most} refer to both noise in the data and errors in prediction and explanation.} $v \in V(G)$ with $f(v)=\classlabel$, there is a rooted pattern $(P_c,r_c) \in S_\classlabel$ and a rooted subgraph isomorphism from $(P_c, r_c)$ to $(G, v)$ and (ii) there is \textit{no} such rooted pattern in $S_\classlabel$ and rooted subgraph isomorphism for \textit{most} $v' \in V(G)$ with $f(v')\neq\classlabel$.
The set $S_\classlabel$ is the underlying \textit{ground truth}.
%
%
%

Let $v \in V(G)$ be a node selected uniformly at random such that $\Phi(v) = c$ for some $c \in C$ and let $(P_c,r_c)$ be a rooted pattern from $S_\classlabel$.
It follows from the assumptions that with a certain probability, a rooted explanation graph $(P_v,v) \in \mathcal{E}_c$ computed for $v$ contains a subgraph $P_v'$ such that $v \in V(P_v')$ and $(P_v',v)$ is a rooted subgraph of $(P_c,r_c)$ (i.e., there is a rooted subgraph isomorphism from $(P_v', v)$ to $(P_c, r_c)$).
Since $P_v'$ is a subgraph of $P_v$ and rooted subgraph isomorphism is used for pattern matching, $(P_v',v)$ contains \textit{less} structural constraints.
Hence, it can be regarded as a \textit{generalization} of 
$(P_v,v)$.\footnote{By generalization we mean the relationship between two rooted patterns in the poset of all rooted patterns defined by rooted subgraph isomorphism, and \textit{not} ``generalization'' from data as used in machine learning.}
%
Thus we need to calculate a set of rooted patterns that generalize a large fraction of the rooted explanation graphs in $\mathcal{E}_c$  
and, in order to avoid redundancy, are \textit{most specific}  with respect to this property.
As mentioned above, 
there are two sources of errors for the explanation patterns computed by the previous module. 
First, with a certain probability, the true class label $f(v)$ is predicted \textit{incorrectly} by $\Phi$ (i.e., $f(v) \neq c = \Phi(v)$). 
Second, there is \textit{no} guarantee that the explanation graph $(P_v,v)$  provides a \textit{genuine} explanation for predicting the class label of $v$ by $\Phi(v)=c$, independently whether or not $\Phi(v)=f(v)$.
Still, 
it is reasonable to assume 
that many of the \textit{individual} explanation patterns computed for $c$ 
have a relatively large overlap with some unknown class patterns in  $S_\classlabel$.
Accordingly, we expect that \textit{most}  \textit{frequent} rooted subgraphs of $\mathcal{E}_c$ computed in the node explanation module are rooted \textit{subgraphs} of some patterns in  $S_\classlabel$.

The above arguments motivate considering the \textit{maximal frequent} rooted subgraphs of $\mathcal{E}_c$ (i.e., which are frequent and all their proper connected supergraphs are infrequent in $\mathcal{E}_c$  w.r.t. $\tau$) (line~\ref{line:maximal_frequent_patterns}). The assumptions imply that they contain at least a part of the structural information assigning a node to class $c$.    
These patterns can be regarded as the \textit{most specific generalizations} of most of the individual explanations in $\mathcal{E}_c$. (
(For more details please see Appendix~\ref{sec:patternextractionmodule}.)
%

After the computation of the maximal frequent rooted subgraphs for all $c \in C$ (lines~\ref{line:start_pattern_extraction}--\ref{line:maximal_frequent_patterns}), a small subset $\mathcal{P}^\top$ of \textit{top} rooted patterns is selected from $\mathcal{P} =\bigcup_{c\in C} \mathcal{P}_c$ (line~\ref{line:select_top_patterns} ). 
More precisely, for each $(P,r) \in \mathcal{P}_c$, 
EEGL evaluates how well $(P,r)$ performs as a classifier for class $c$ by computing its F1-score on the training set $T$. 
The module then returns the top $d'$ patterns with the highest F1-scores in a round-robin fashion from $\mathcal{P}$, 
where $d' = \min\{d,|\mathcal{P}|\}$. 

\textbf{Feature Annotation Module (line~\ref{line:feature_annotation})} Using 
the rooted patterns $(P_1,r_1),\ldots,(P_{d'},r_{d'})$ returned by the pattern extraction module, 
in this module we update 
$\featurematrix$ for the next iteration of the main loop 
by setting $\featurevector_v[j]$ to $1$ if there is a rooted subgraph isomorphism from $(P_j,r_j)$ to $(G,v)$; otherwise to $0$, for all $v\in V(G)$ and $j \in [d']$. 
If $d' < d$, the last $d-d'$ entries in $\featurevector_i$ are set to some constant.

\section{Experimental Evaluation}
\label{sec:experiments}

In this section we present our experimental results
evaluating the predictive performance of {\EEGL} on synthetic and real-world graphs 
and address the questions formulated in Sect.~\ref{sec:introduction}.
All learning tasks are themselves or are reduced to node classification tasks.
The use of synthetic data is also justified by
~\cite{Maekawa22}, discussing the state of the art on node classification benchmarks. 
The focus of~\cite{Maekawa22} is on parameters such as class size distributions, 
and edge densities between classes (corresponding to heterophily versus homophily).
In contrast, we are interested in the effect of  \emph{structural complexity}.
The experiments are grouped into three blocks, depending on the relationships between the graphs' 1-WL and target classes, determined by the graph structures and the class labels, respectively.  

\subsection{Experimental Setup and Runtimes}
Recall from Section~\ref{sec:EEGL} that the input graph $G$ with $n$ nodes is associated with a feature matrix 
of size $n \times d$. 
For each graph $G$ used in the experimental evaluation, we  carried out different experiments with GCN~\citep{kipf17-gcn}, 
\shadowgnn~\cite{Zeng_etal/2021}, and our {\EEGL} system. 
The GCN experiments use precomputed feature vectors. 
In particular, in the \textit{label encoding} setting, feature vectors are essentially  one-hot encodings of the node labels. 
In the \textit{adversarial} setting, feature annotation is done with a precomputed set of adversarial rooted patterns constructed for the ground truth patterns.
In the \textit{randomized} setting, feature annotation is done using precomputed random numbers from $[0,1]$. 
Motivation for choosing these settings and discussion of the experiments are given in Sections~\ref{sec:old_motifs}--\ref{sec:fullerenes}.

\textbf{Parameters} Regarding the parameters, we used different values for $d$, depending on the number of target classes. 
Our realization of {\EEGL} uses {\Gaston}~\citep{Nijssen/Kok/04} to generate the candidate frequent patterns. This subroutine has 
a frequency threshold  parameter $\tau$.
We used $\tau = 0.7$ for the datasets in Sections~\ref{sec:old_motifs}, \ref{sec:new_motifs} and $\tau = 0.1$ for the fullerenes in Sect.~\ref{sec:fullerenes}.
The GCN model and training parameters (size of hidden layers, dropout rate, number of epochs, learning rate, and weight decay) were chosen by hyperparameter grid search.
Recall from Sect.~\ref{sec:EEGL} that in case of vanilla initialization, $X$ is set initially to the matrix of ones.
The learning step in the initial round can therefore be regarded as learning a vanilla GNN because the feature matrix provides \textit{no} information. 
It is therefore not counted as an {\EEGL} round and will be referred to as round~0 or R0. 
Using this terminology, we carried out at least one iteration with {\EEGL} in all experiments.
The same applies to {\EEGL} with random initialization.
Finally, for all experiments with GCN, {\shadowgnn}, and {\EEGL}, we used $k$-fold cross validation (CV) for $k= 5$ or $10$, depending on graph size. 


\textbf{Implementation Details} Experiments showed that the performance of EEGL framework is highly influenced by the choice of the classifier and the explainer implementations. We noticed that for the synthetic dataset derived by addition of subgraph motifs to Barab\'asi-Albert random graph, the GNNExplainer reference implementation~\cite{ying19-gnn_explainer} performed better. This could be attributed to the fined-tuned denoising step and other domain specific architectural nuances in the reference implementation that were adapted for a better performance on the house dataset and its derivatives. For all other datasets, a standard out-of-the-box combination of GCN + GNNExplainer from the open source PyG~\cite{Fey/Lenssen/2019} library was sufficient to yield very high accuracy. We used Glasgow Subgraph Solver~\cite{Glasgow/2020}, a high-performance subgraph isomorphism solver, for deciding rooted subgraph isomorphism. Most of the experiments were run on a machine(s) with the following architecture: \textbf{processor}: 72 cores of Intel(R) Xeon(R) Gold 6240 CPU @ 2.60GHz, \textbf{memory}: 392G, \textbf{graphics processor}: Nvidia Tesla V100-PCIE-32GB. 


\textbf{Runtime Analysis} 
On the largest graphs in the experiments with nearly 1,000 nodes, EEGL spent an average of 17 min. (1,044.66 sec.) for one iteration and for a single fold, with the following distribution among the modules (also in percentage):  
11.54  sec. (1.1\%) on the learning,  
444.75 sec. (42.6\%) on the explainer,
456.07 sec. (43.6\%) on the pattern extraction, and 
132.30 sec. (12.7\%) on the feature annotation modules.
Since EEGL computes everything from scratch in each iteration, the runtime grows \textit{linearly} with $K$ (number of iterations in Alg.~\ref{alg:EEGL}).
We note that {\EEGL} was faster on all other graphs used in the experiments.

\subsection{Graphs with Low 1-WL Symmetry}
\label{sec:old_motifs}

\begin{figure*}[t]
  \centering
  \hspace*{-1cm}
  \begin{subfigure}{0.2\linewidth}
    \includegraphics[width=0.9\linewidth, left]{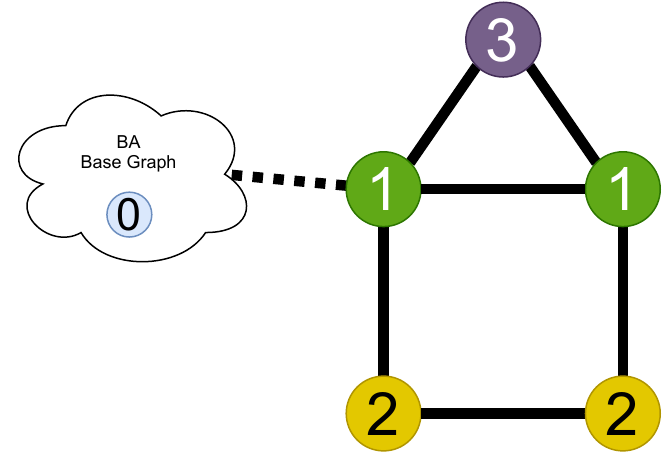}
    \caption{\centering $M_1$}
    \label{fig:m1}
  \end{subfigure}\hspace*{-0.6cm}
  \begin{subfigure}{0.24\linewidth}
    \centering
    \includegraphics[width=0.87\linewidth]{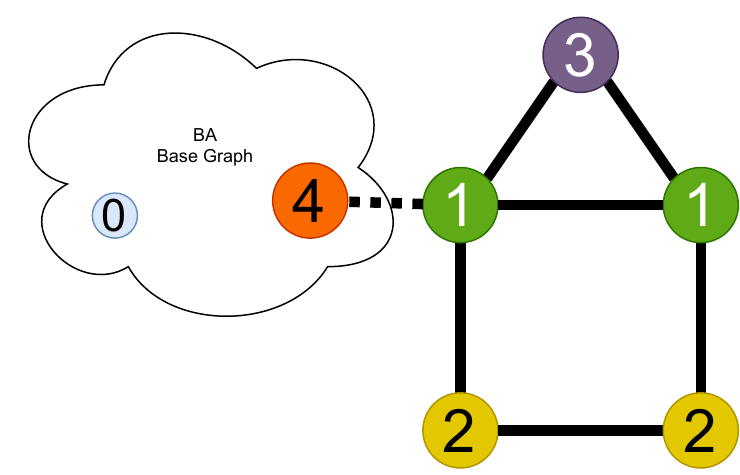}
    \caption{$M_1'$}
    \label{fig:m1p}
  \end{subfigure} \hspace*{-1em}
  \hspace*{0.75cm}
  \begin{subfigure}{0.26\textwidth}
    \centering
    \includegraphics[width=1\linewidth]{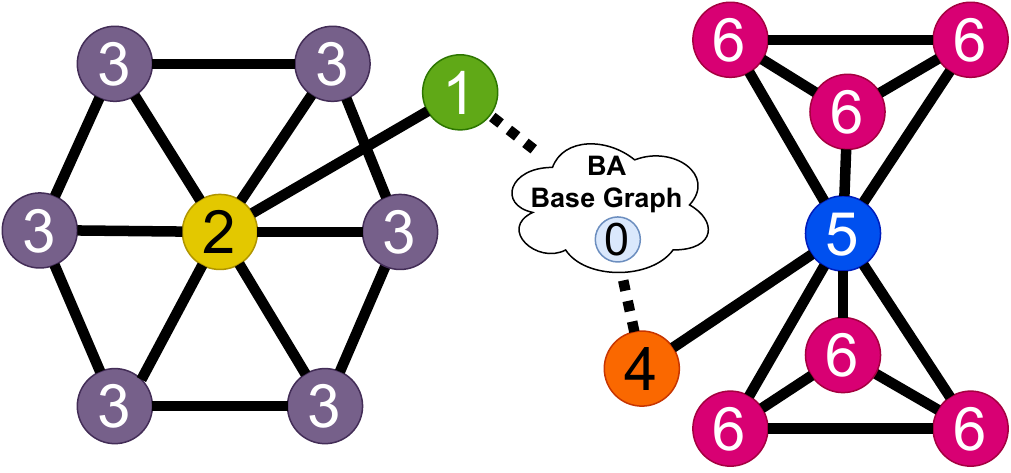}
    \caption{\centering $M_2$} 
    \label{fig:m2}
  \end{subfigure}\hspace{0.75cm}
  \begin{subfigure}{0.24\textwidth}
    \centering
    \includegraphics[width=1\linewidth]{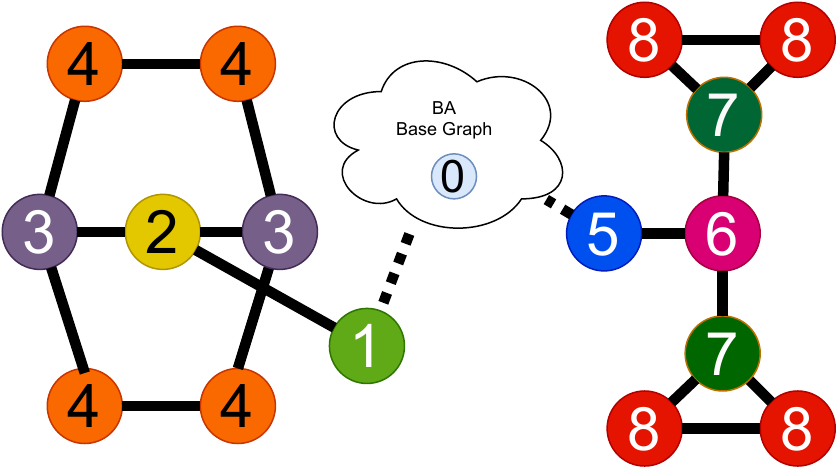}
    \caption{\centering $M_2'$}
    \label{fig:m2p}
  \end{subfigure} \hfill
  \caption{The motifs used in the dataset generation: The ``house'' motif (a) and its variant (b), and motif pairs with 1-WL indistinguishable nodes (c,d). }
  \label{fig:motifs}
  \vspace{-1em}
\end{figure*}

In this section we study the questions formulated in Sect.~\ref{sec:introduction} for graphs for which all 1-WL classes contain relatively few elements only, compared to the graphs' size. 
Since nodes belonging to different 1-WL classes are 1-WL distinguishable, such graphs have \textit{low} 1-WL symmetry. 
They have \textit{no} 1-WL symmetry if all 1-WL classes are singletons.  
In addition to this property, for all target classes $c$, the set of nodes in $c$ is formed by the union of a family of 1-WL classes. 
For the experiments, we adopt the synthetic model of random graphs with motifs attached. 
This model, introduced in \cite{ying19-gnn_explainer}, has become a standard benchmark for GNN. 

\textbf{Datasets}
Graphs considered in this section are constructed as follows: 
Generate a Barab\'asi-Albert (BA) random base graph~\citep{barabasi99-emergence} and attach copies of 
small graph motifs to $m$ random nodes.
The motifs are shown in Fig.~\ref{fig:motifs}, together with specifying how to attach them to the BA graph. 
For all motifs we used the same node and motif numbers as in \citep{ying19-gnn_explainer}, i.e., $n = 300$ and $m = 80$.
For $M_2$ and $M_2'$, 
select one of the two motifs uniformly at random for each of the $m$ copies 
and assign then the class labels to the nodes. The class labels of the motifs' nodes are indicated in Fig.~\ref{fig:motifs} and the nodes of the BA graph are all labeled by $0$, except for the node with label 4 for $M_1'$.
Finally, add a small amount of \textit{structural noise} to the motifs via a small set of random edges.
The graphs obtained in this way for $M_1,\ldots,M_2'$ in Fig.~\ref{fig:motifs} are denoted by $\GM{1},\ldots,\GMp{2}$, respectively.

Note that for each motif $M$ in Fig.~\ref{fig:motifs}, all nodes of $G(M)$ with label $\ell$ can be distinguished from the nodes with label $\ell'$ by some characteristic rooted pattern(s), for all $\ell > \ell' > 0$. 
For the nodes with label $0$, we are not aware of a small set of such characteristic patterns.
Class 0 is handled in the same way as all other classes.

The following considerations motivated our choice of the motifs. 
Regarding $M_1$ (Fig.~\ref{fig:m1}) used in {\GNNExplainer}~\citep{ying19-gnn_explainer}, the three labels indicate its 1-WL classes.
Its attachment to the base graph 
\textit{breaks}, however, the 1-WL symmetry.  
In fact, $\GM{1}$ in the experiments had \textit{no} 1-WL symmetry (i.e., all of its 1-WL classes were singletons).
%
The only difference between $\GMp{1}$ (see Fig.~\ref{fig:m1p} for $M_1'$) and $\GM{1}$ is that the base graph's nodes are split into two classes by assigning label $4$ to the attachment nodes. 
This kind of ``merged'' node classes are more challenging for vanilla GNNs.
For both of the last two motif pairs in Fig.~\ref{fig:motifs} we have that certain nodes within a pair are indistinguishable by  1-WL (e.g., the nodes with label $3$ from those with label $6$ in $M_2$ in Fig.~\ref{fig:m2}) and hence 
by vanilla GNNs as well.
For $\GM{2}$ and $\GMp{2}$, however, all base graph nodes belonged to a singleton 1-WL class and all motif nodes within a 1-WL class had the same class label and were in the same motif occurrence.  

\textbf{Experimental Results and Analysis}
\label{sec:exper}
\begin{table}[t]
\begin{footnotesize}
\begin{tabular}{lcccc}
\toprule
              &  $\GM{1}$ & $\GMp{1}$ & $\GM{2}$ & $\GMp{2}$\\
\midrule
Label Encoded (LE) & 100.00$\pm$0.00 & 100.00$\pm$0.00 & 99.58$\pm$0.51 & 99.89$\pm$0.34  \\ \hline
Adversarial (A) & 97.71$\pm$1.14 & 80.50$\pm$4.59 & 79.45$\pm$4.39 & 71.51$\pm$5.95 \\
Random (R) & 96.38$\pm$2.46 & 93.23$\pm$3.79 & 95.62$\pm$4.44  & 79.57$\pm$3.98 \\
{\shadowgnn}       &  \textbf{100.00$\pm$0.00} &  96.29$\pm$4.89 & 73.09$\pm$7.07 & 87.65$\pm$9.00 \\ 
{\EEGL} Round-0 (R0) & 98.56$\pm$1.83 & 90.82$\pm$4.12 & 61.53$\pm$3.39 & 56.81$\pm$3.40 \\
EEGL Round-1 (R1) & 99.86$\pm$0.42 & \textbf{99.29$\pm$0.71} & \textbf{99.45$\pm$0.55} & \textbf{99.36$\pm$0.69}  \\
\bottomrule
\end{tabular}
\caption{Average weighted F1-score results in percentage ($\text{mean}\pm\text{SD}$) obtained with 10-fold CV for the motifs in Fig.~\ref{fig:motifs} with the label encoding (LE), adversarial (A), random (R) settings, {\shadowgnn}, vanilla GNN (R0), and rounds~1 of {\EEGL} (R1).
}
\label{table:results}
\end{footnotesize}
\vspace*{-3em}
\end{table}
The results are presented in Table~\ref{table:results}. 
The rows correspond to the feature matrix definitions in 
GNN (LE, A, R), {\shadowgnn}~\cite{Zeng_etal/2021}, vanilla GNN (R0), and the first iteration of {\EEGL} (R1).
We used 10-fold CV and report the mean weighted F1-scores (percentage), together with the standard deviations. 
%
As expected, GNN with the label encoding setting (LE) 
consistently achieved a result close to 100\%. Thus, values in row LE should be regarded as \textit{upper bounds} on performance achievable by GNN.
%
Below we answer the questions formulated in Sect.~\ref{sec:introduction} for graphs with low 1-WL symmetry. 
The quantitative experimental results are accompanied with an \textit{analysis}. 
We peek under the hood of {\EEGL} to understand its \textit{training dynamics}. 

\begin{figure}[t]
  \centering
  \includegraphics[width=\linewidth]{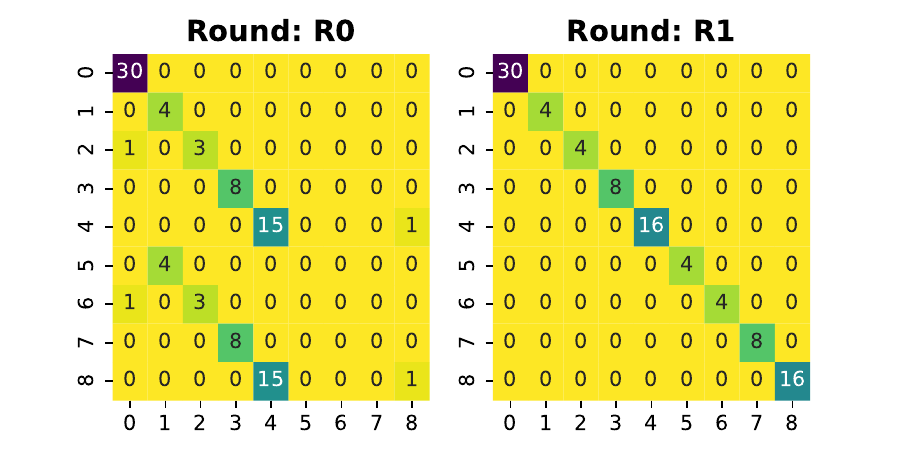}
  \vspace{-2em}
  \caption{Example of the confusion matrices for a fold of $\GMp{2}$.}  \label{fig:cm_m2p}
  \vspace{-1em}
\end{figure}

\textbf{Answer to Q1} We answer this question 
by comparing the results with vanilla GNN (R0) to those with {\EEGL}'s first iteration (R1) (see Table~\ref{table:results}).
The order of the graphs in Table~\ref{table:results} follows their difficulty for vanilla GNN. 
While the results are inconclusive for $\GM{1}$ (R0 and R1 are both close to 100\%), there is an improvement for all other graphs. 
In particular, {\EEGL} achieved in one iteration a considerable improvement of around 38\% and 43\% on $\GM{2}$ and $\GMp{2}$, respectively.
For example, out of the 94 test nodes, vanilla GNN (R0) classified only 61 correctly (see the confusion matrices in Fig.~\ref{fig:cm_m2p} for the test examples for one of the 10 folds for $\GMp{2}$).
The number of correctly classified nodes increases to 94 (100\%) in one round of {\EEGL} (R1).
This and other results of this section indicate that
\textit{{\EEGL} is capable of iterative self-improvement from explanations already in one iteration}, giving an affirmative answer to Q1. 


\textbf{Answer to Q2} 
As noted earlier, LE (label encoding) provides an \textit{upper bound} for achievable prediction, and it achieved almost perfect prediction for each variant.
The ``adversarial'' (A) feature initialization addresses the issue that any set of subgraph features is expected to improve upon the prediction performance of GNN ~\citep{barcelo21-graph_params}. Comparing the improvement of EEGL with other feature initializations seems to be nontrivial. One possibility would be to consider random subgraph features, but it is not clear what would be the relevant notion of a random graph here. We consider a weakest possible formulation, and try to show that EEGL brings more improvement than \emph{some} arbitrary feature initialization. This, however, includes cases when the subgraph features do not occur at all, and thus feature initialization trivializes. Therefore we attempt to find an ``unhelpful'' but nontrivial subgraph initialization. The approach is to use subgraph features which do not occur  in the motifs, but may occur in the base graph. As the base graph is random, the features are expected to be of limited help. Note that motif nodes get a trivial initialization, but the nontrivial initialization of the base graph nodes may have an effect on them.
Similarly to EEGL, these features are problem dependent as well. The experiments show that indeed, \textit{EEGL (R1) is better than A} (see Table~\ref{table:results}).
Randomized feature initialization has been shown to be powerful in a theoretical sense~\cite{Abbou21,Dasou}. Randomization extends GNNs in a different direction than EEGL. 
The comparison shows that \textit{{\EEGL} (R1) outperforms randomization (R)} already in one iteration on all graphs. 
Finally, while the difference between {\EEGL} (R1) and {\shadowgnn}~\cite{Zeng_etal/2021} is negligible on $\GM{1}$ and $\GMp{1}$, {\EEGL} performs much better on $\GM{2}$ and $\GMp{2}$. 
%
In summary, the results on synthetic graphs with low 1-WL symmetry show that \textit{{\EEGL} outperforms other node feature definitions and {\shadowgnn} in predictive performance} already in one iteration, giving an affirmative answer to Q2 for this kind of graphs.

\begin{figure}[t]
  \centering
  \includegraphics[width=\linewidth,height=1.2cm]{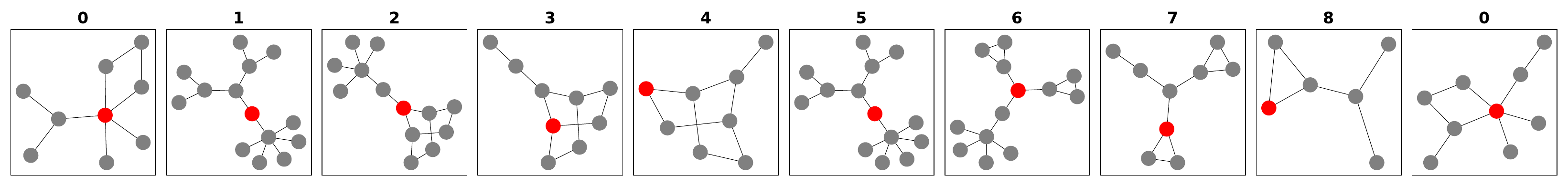}\vspace{-0.3cm}
  \caption{The $d=10$ maximal frequent subgraphs extracted by {\EEGL} in the first (R0 $\to$ R1) iteration for a fold of $\GMp{2}$. Class labels indicated on top. There are two patterns for label 0.}
  \label{fig:fsg_m2p}
  \vspace{-1em}
\end{figure}

\textbf{Answer to Q3} 
Consider again $M_2'$ (Fig.~\ref{fig:m2p}) and the confusion matrices in Fig.~\ref{fig:cm_m2p} (see Appendix~\ref{sec:detailedresults} for the detailed results on $\GMp{2}$).
Prediction mistakes made by vanilla GNN (R0) indicates that it is sometimes unable to distinguish nodes belonging to different 1-WL classes in $\GMp{2}$, but to the same one locally, i.e., in $M_2'$. As an example, for the nodes in the 1-WL indistinguishable classes 4 and 8 (cf. Fig.~\ref{fig:m2p}) vanilla GNN predicts 1 node of class 4 by class 8 and conversely, 15 nodes of class 8 by 4. 
Out of 33 mistakes, 31 are of this type; the remaining two arise from classifying motif nodes with label 0.
The picture for the first iteration of {\EEGL} is different (see R1 in Fig.~\ref{fig:cm_m2p}).
Consider, for example, the label pair 4--8. 
The top 10 frequent patterns extracted in round~0 are given in Fig.~\ref{fig:fsg_m2p} (same fold as in Fig.~\ref{fig:cm_m2p}). 
One can check that the patterns for labels 4 and 8 are genuine in the sense that they distinguish nodes in these classes. 
The situation for label pairs 2--6 and 3--7 is similar.
This is, however, not the case for 1--5. Still, {\EEGL} (R1) can correct all mistakes on nodes of class 5. We speculate that this is because the neighbors of the nodes of the two classes have different feature vectors. 
In summary, our answer to Q3 is that the results suggest that \textit{{\EEGL} has an excellent performance on learning over graphs having low 1-WL symmetry and target classes are defined by unions of 1-WL classes}. 

\subsection{Graphs with High 1-WL Symmetry}
\label{sec:new_motifs}

\begin{figure*}[t]
	\centering
	\begin{subfigure}[c]{0.2\textwidth}
		\centering
		\includegraphics[width=1\textwidth]{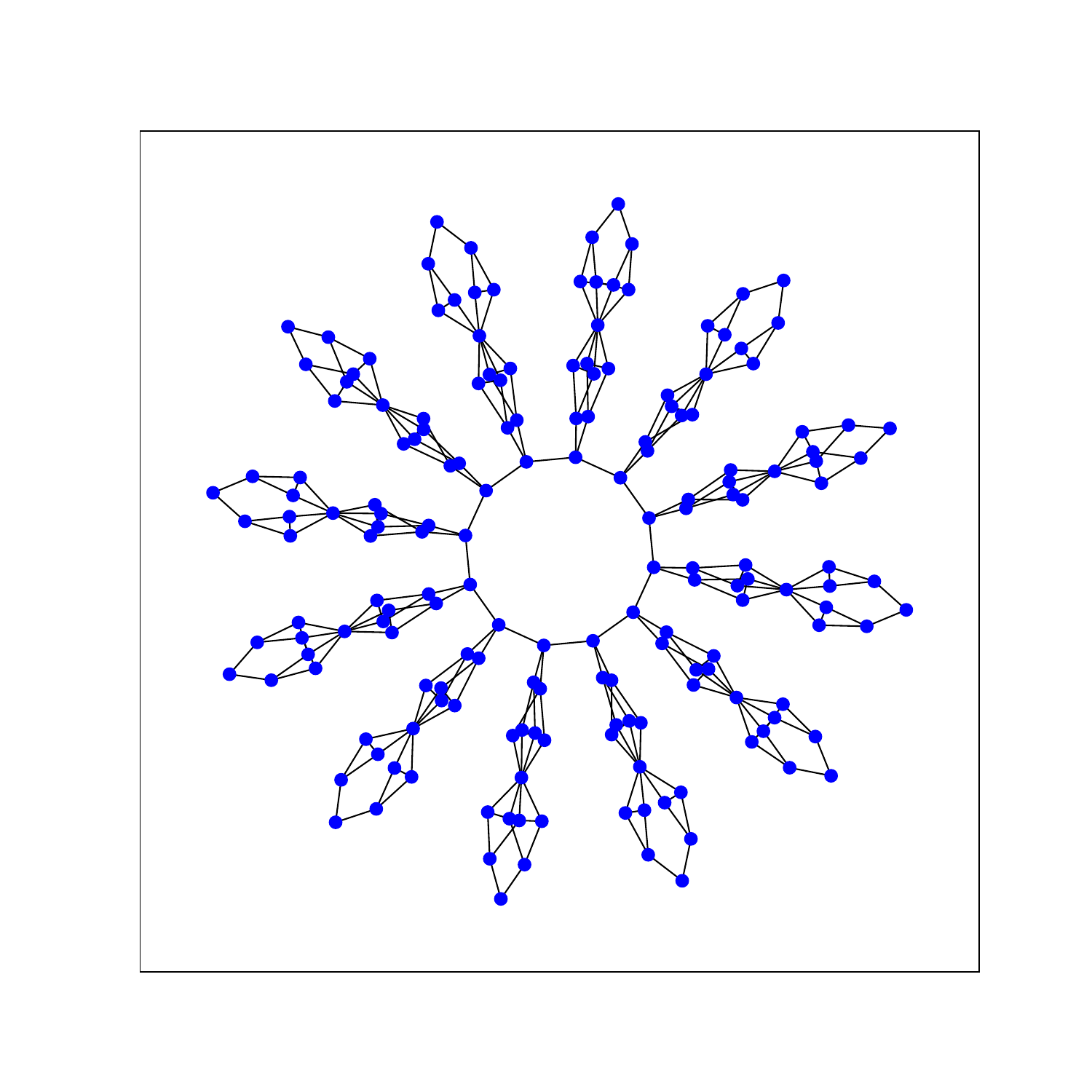}
		\caption{\centering $G_{180}$}
		\label{fig:G180}
	\end{subfigure} \hfill
	\begin{subfigure}[c]{0.09\textwidth}
		\centering
		\includegraphics[width=1\textwidth]{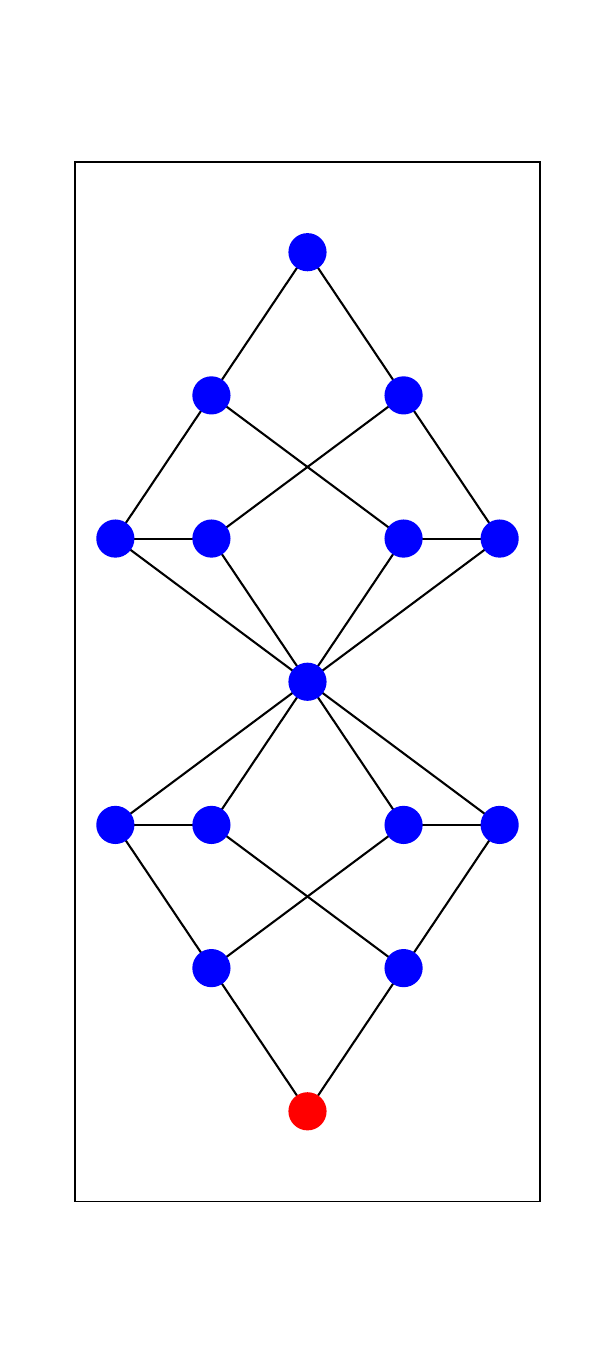}
		\caption{\centering $M_0$}
		\label{fig:AA}
	\end{subfigure} \hfill
	\begin{subfigure}[c]{0.09\textwidth}
		\includegraphics[width=1\textwidth]{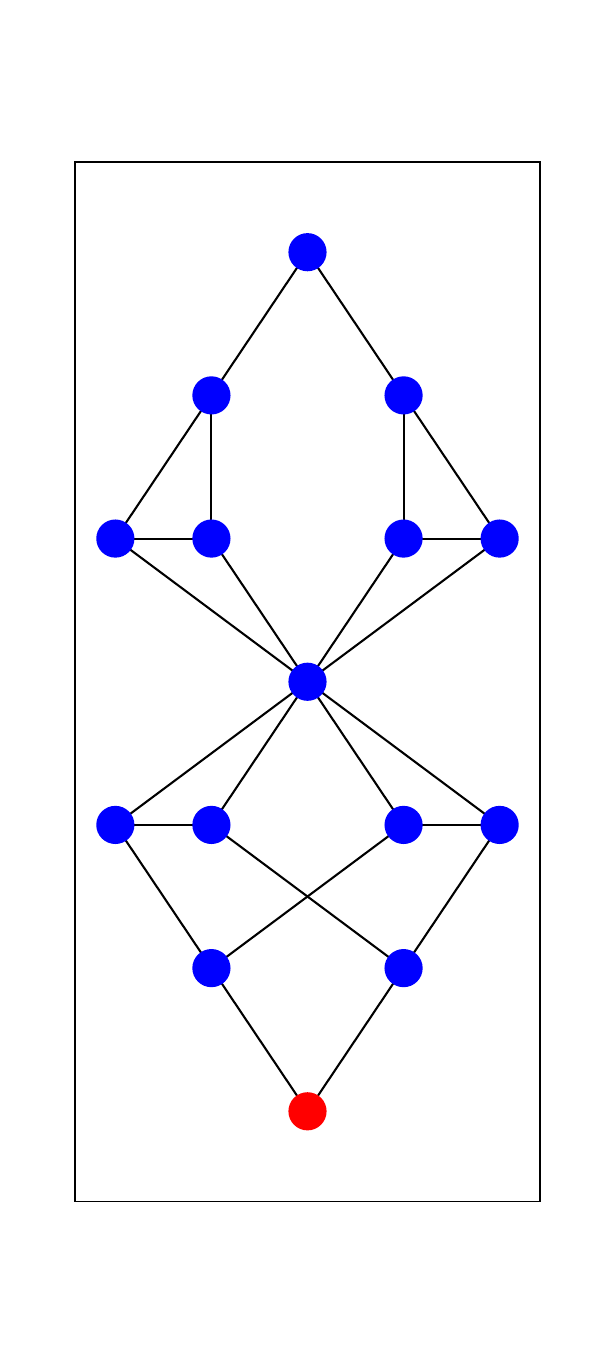}
		\caption{\centering $M_1$}
		\label{fig:AB}%
	\end{subfigure}  \hfill
	\begin{subfigure}[c]{0.09\textwidth}
	\includegraphics[width=1\textwidth]{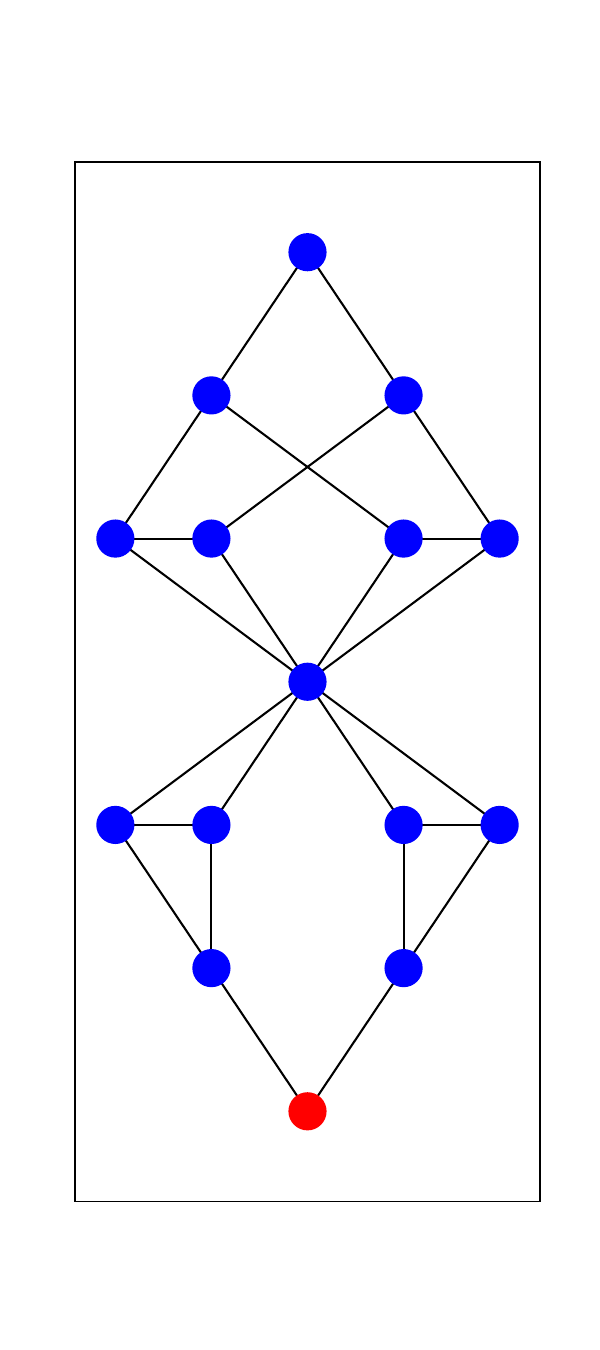}
	\caption{\centering $M_2$}
	\label{fig:BA}%
	\end{subfigure} \hfill
	\begin{subfigure}[c]{0.09\textwidth}
	\includegraphics[width=1\textwidth]{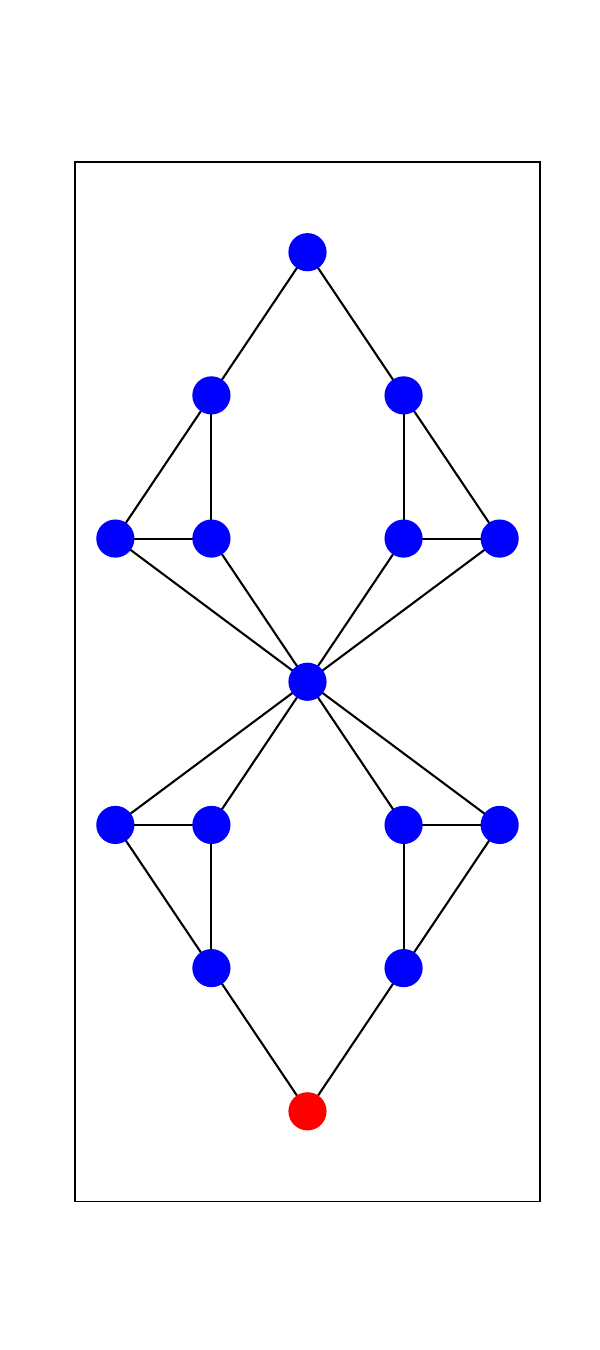}
	\caption{\centering $M_3$}
	\label{fig:BB}%
	\end{subfigure}
		\begin{subfigure}[c]{0.1\textwidth}
		\centering
		\includegraphics[width=1\textwidth]{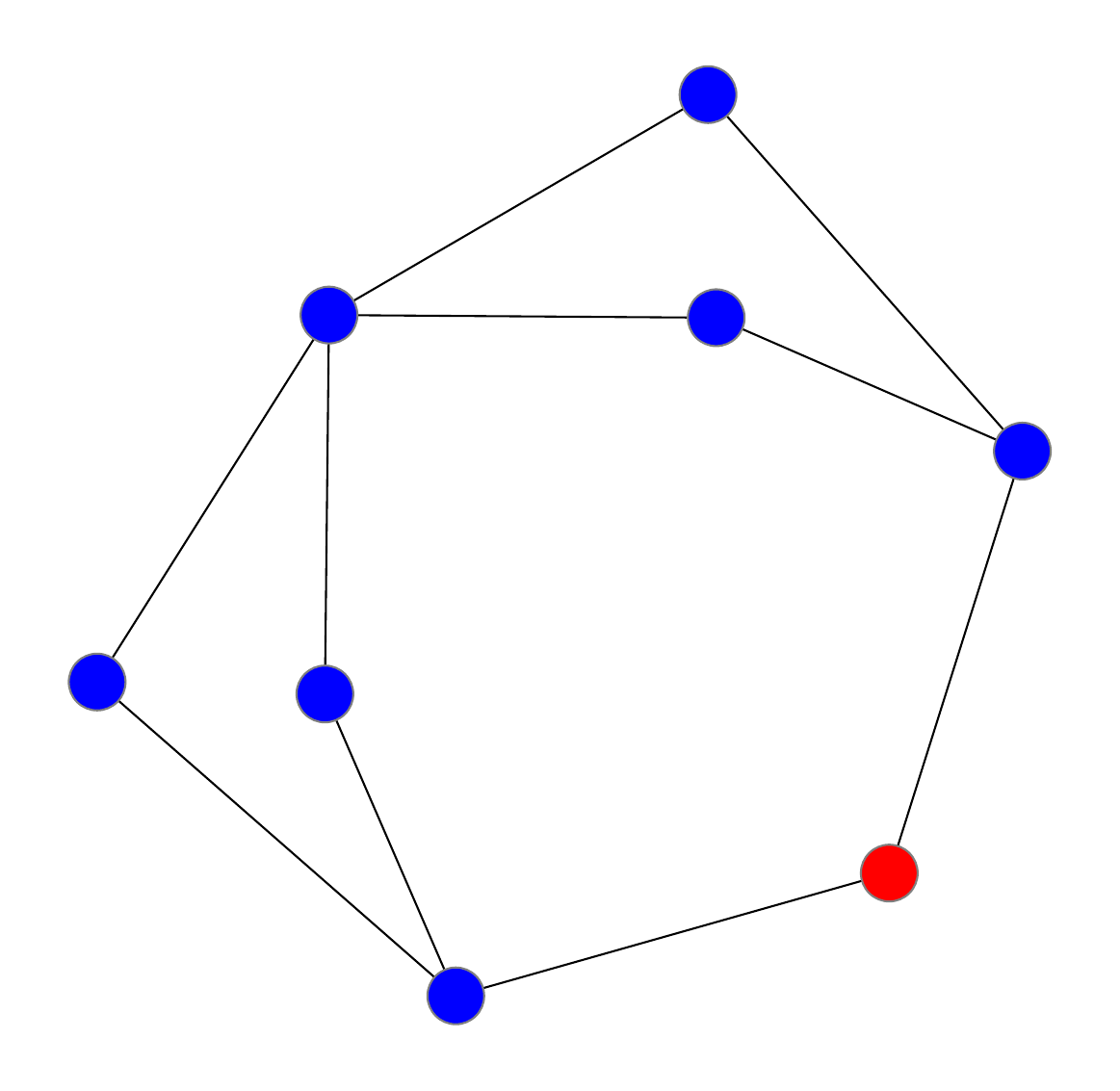}
		\caption{\centering $(P_1, r_1)$}
		\label{fig:G180_motif_1}
	\end{subfigure} \hfill
	\begin{subfigure}[c]{0.1\textwidth}
		\centering
		\includegraphics[width=1\textwidth]{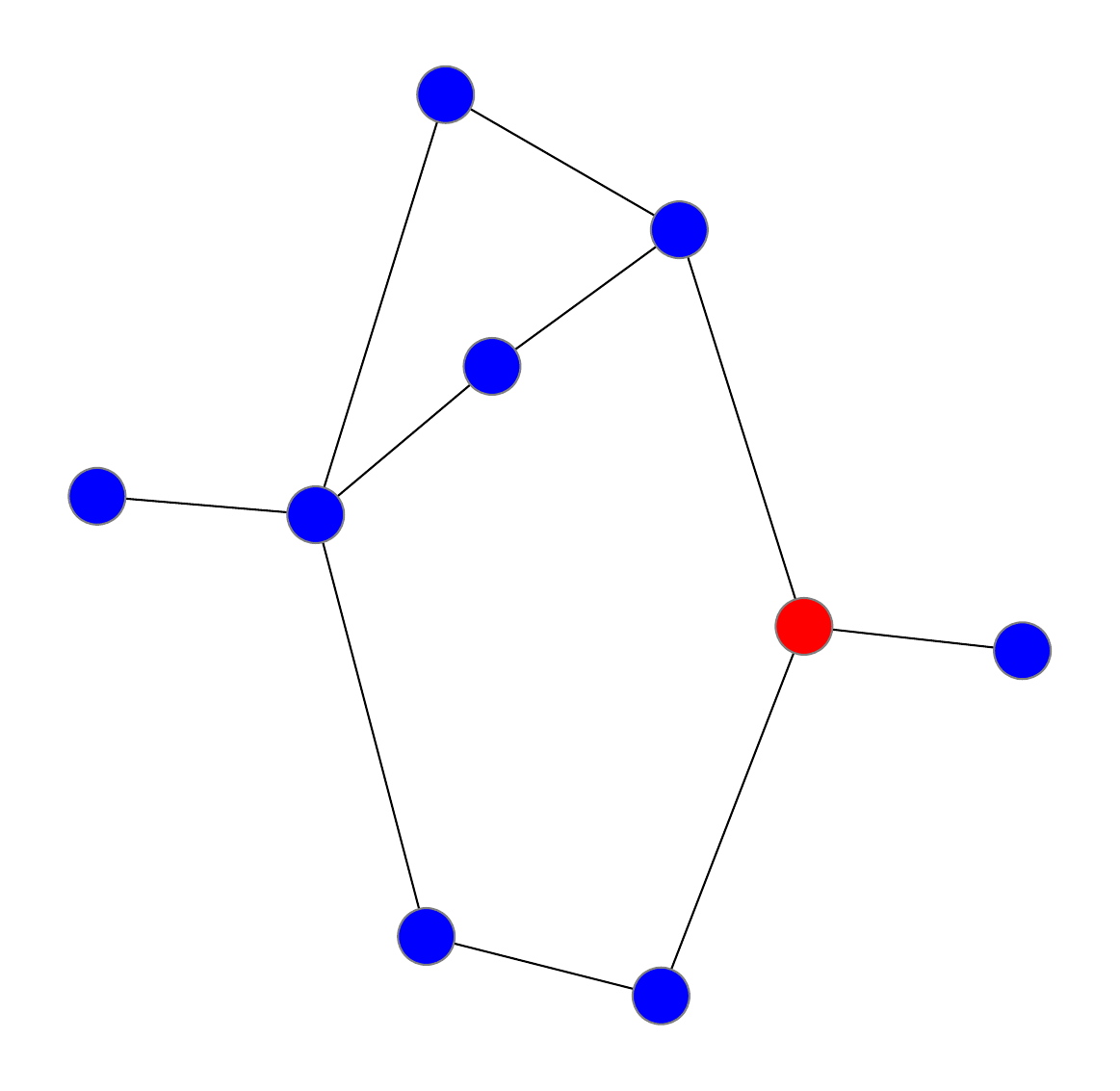}
		\caption{\centering $(P_2,r_2)$}
		\label{fig:G180_motif_2}
	\end{subfigure} \hfill
	\begin{subfigure}[c]{0.1\textwidth}
		\includegraphics[width=1\textwidth]{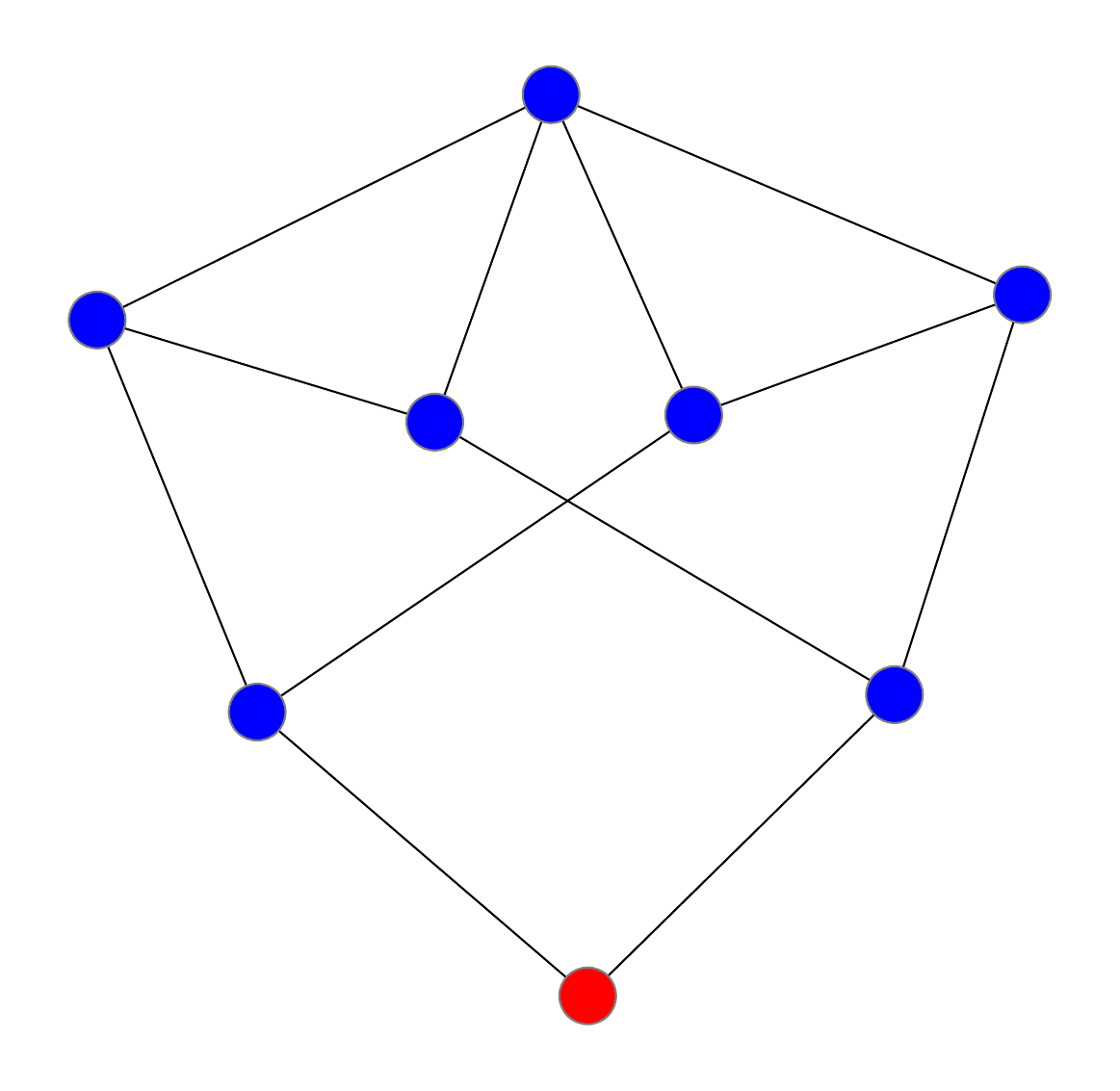}
		\caption{\centering $(P_3,r_3)$}
		\label{fig:G180_motif_3}%
	\end{subfigure}  \hfill

	\begin{subfigure}[c]{0.2\textwidth}
		\centering
		\includegraphics[width=1.13\textwidth]{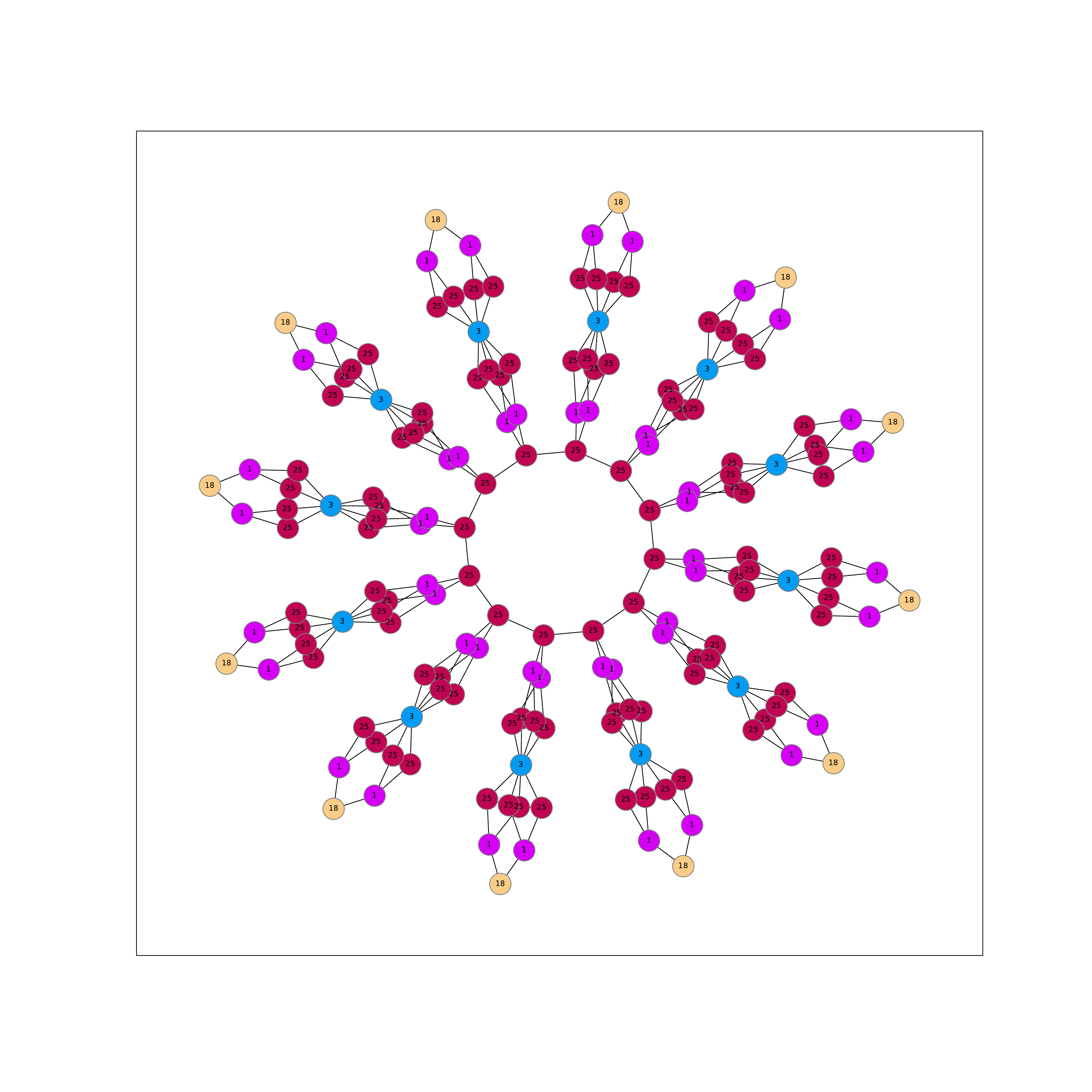}
		\caption{\centering R0}
		\label{fig:G180_r0}
	\end{subfigure} \hfill
	\begin{subfigure}[c]{0.2\textwidth}
		\centering
		\includegraphics[width=1.13\textwidth]{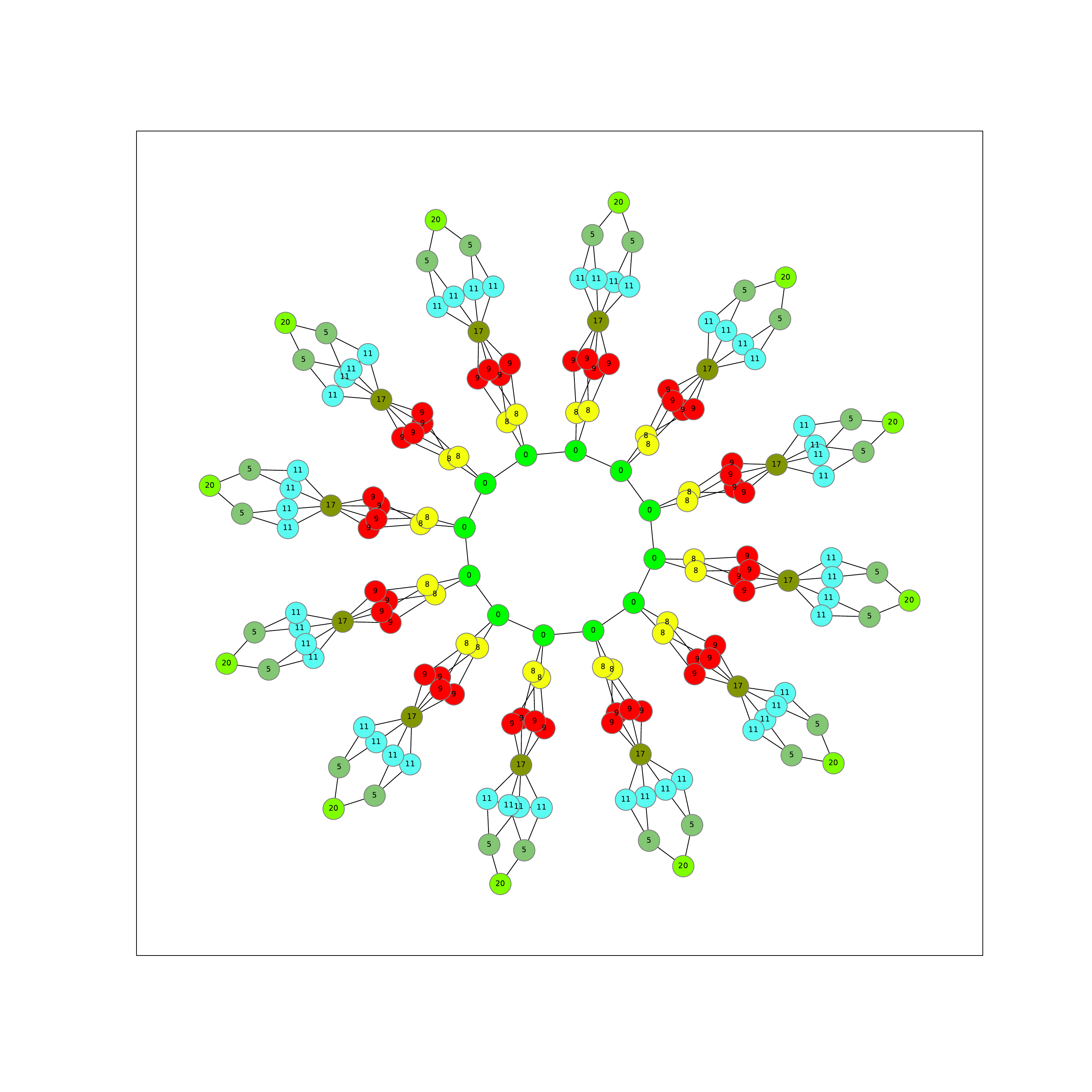}
		\caption{R1}
		\label{fig:G180_r1}
	\end{subfigure} \hfill
	\begin{subfigure}[c]{0.2\textwidth}
		\includegraphics[width=1.13\textwidth]{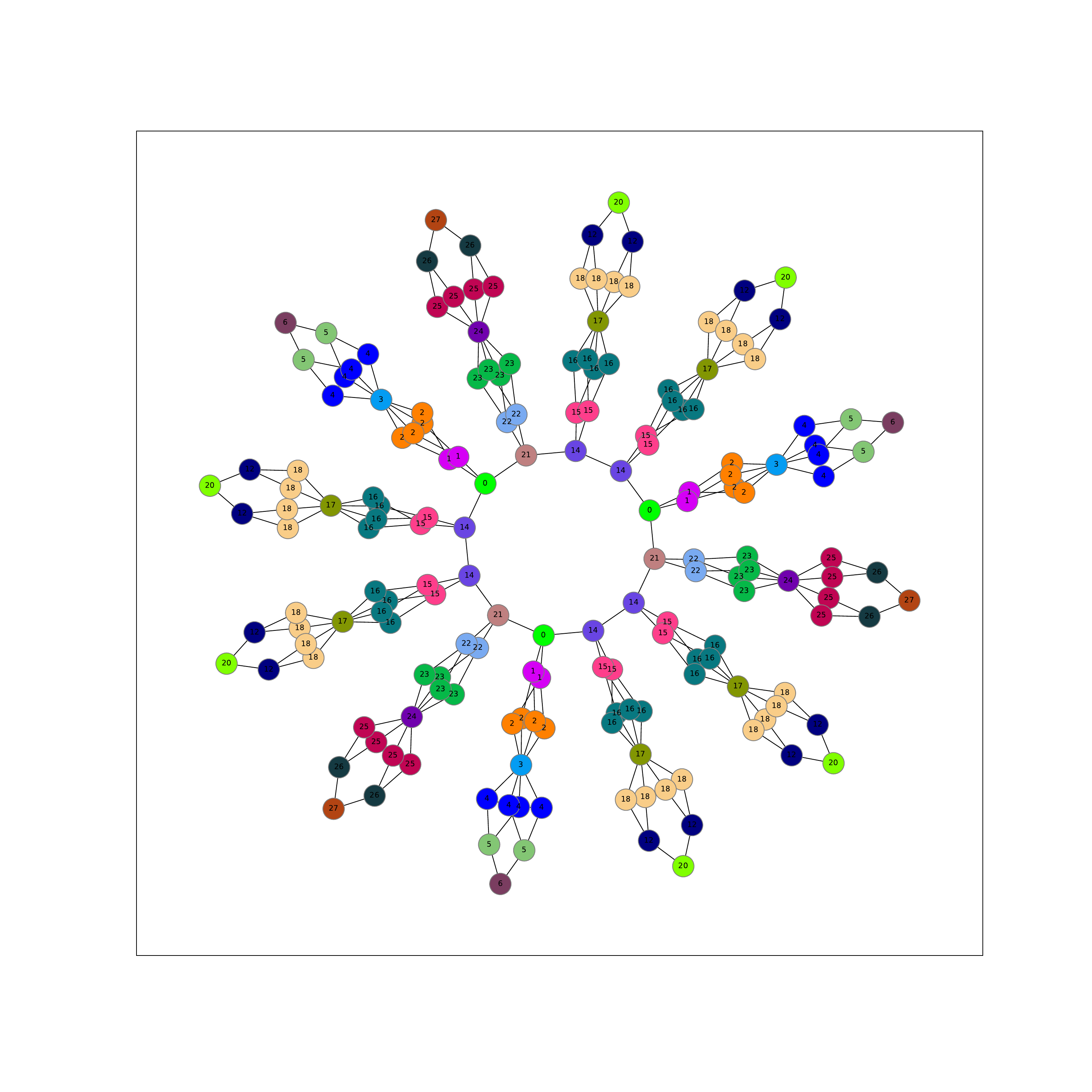}
		\caption{R2}
		\label{fig:G180_r2}%
	\end{subfigure} \hfill
	\begin{subfigure}[c]{0.2\textwidth}
	\includegraphics[width=1.13\textwidth]{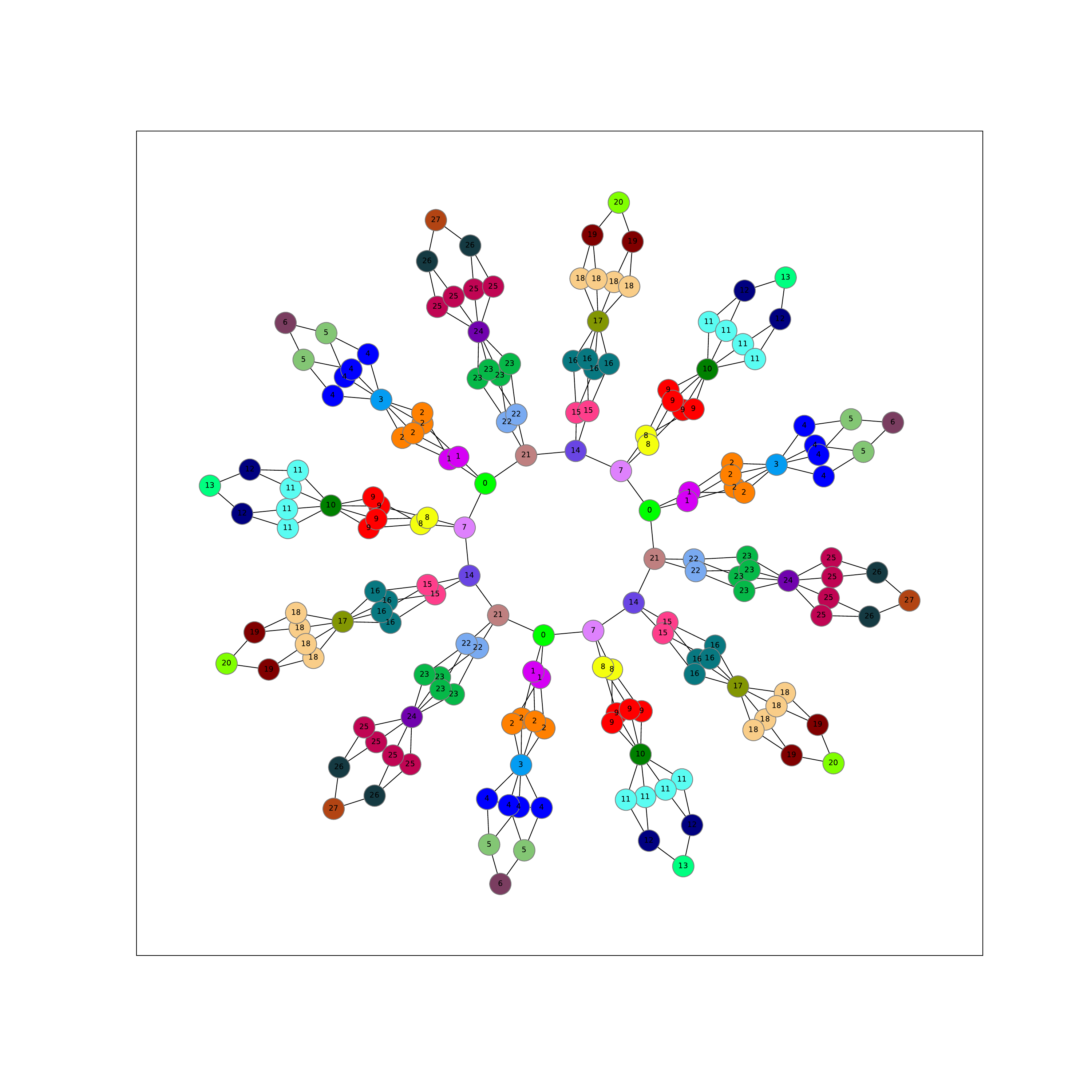}
	\caption{R3}
	\label{fig:G180_r3}%
\end{subfigure}
\caption{Graph $G_{180}$ (a) obtained by attaching the four motifs in (b)--(e) via their bottom nodes (color red) in a rotating manner to the cycle of length 12 in the middle. The 28 target classes, denoted by the colors, are defined by the orbits.
Top frequent patterns extracted by {\EEGL} from the explanations in rounds 2 (f and g) and 3 (h) for $G_{180}$ in Fig.~\ref{fig:G180} with their roots $r_1,r_2,r_3$ marked by red.
The class predictions obtained by {\EEGL} with vanilla initialization is given in (i)--(l). Each of the 28 colors denotes the same class in (i)--(l).}
	\label{fig:M3}
\end{figure*}


\begin{table}[t]
	\begin{footnotesize}
		\begin{tabular}{rll}
			\hline
			LE & \multicolumn{2}{c}{100.00$\pm$0.00} \\ \hline
			{\shadowgnn} & \multicolumn{2}{c}{35.56$\pm$6.95} \\ \hline
			{\EEGL} &  \multicolumn{1}{c}{vanilla} & \multicolumn{1}{c}{random} \\
			R0 & 10.00$\pm$0.00  & 79.78$\pm$6.57\\
			R1 & 56.96$\pm$34.70 & 99.26$\pm$1.48\\
			R2 & 97.78$\pm$2.96 & 100$\pm$0.00 \\
			R3 & 100.00$\pm$0.00 & 100.0$\pm$0.0 \\
			\bottomrule
		\end{tabular}
		\caption{Average weighted F1-score results with SD in percentage obtained for $G_{180}$ with 5-fold CV with the label encoding (LE) setting, with {\shadowgnn}~\cite{Zeng_etal/2021}, vanilla and random initialization (R0), and for three iterations of {\EEGL} (R1, R2, R3).}
		\label{table:G180}
	\end{footnotesize}
        \vspace{-1cm}
\end{table}%


In this section we study a graph of type (ii) in Section~\ref{sec:introduction}, the case opposite to the one in Sect.~\ref{sec:old_motifs}. That is, the graph has a few 1-WL classes compared to its size, i.e., it has \textit{high} 1-WL symmetry, and the equivalence relation defined by the target classes is a refinement of that induced by 1-WL.

\textbf{Dataset} The graph $G_{180}$, given in Fig.~\ref{fig:G180}, 
has 180 nodes and is obtained by attaching the four motifs in Figs.~\ref{fig:AA}--\ref{fig:BB} via their bottom nodes (red) to a cycle of length $12$, denoted by $C_{12}$ (see the middle of Fig.~\ref{fig:G180}). 
More precisely, denoting the nodes of $C_{12}$  by $v_0,\ldots,v_{11}$, $v_i$ is the bottom node of a copy of motif $M_{k}$ for $i \equiv k \pmod{4}$, for every $i = 0,\ldots,11$.
There is no rooted isomorphism between any two motifs in Figs.~\ref{fig:AA}--\ref{fig:BB} mapping their  bottom (i.e., red) nodes to each other.
As a result, $G_{180}$ has 28 orbits, denoted by the colors in Fig.~\ref{fig:G180_r3}.  
These are the target classes. 
Two nodes of $G_{180}$ belong to the same 1-WL class iff they have the same distance from $C_{12}$.
Thus, the target classes define a proper refinement of the 1-WL classes. 

\textbf{Experimental Results and Analysis} 
The mean weighted F1-score results with standard deviations obtained  for $G_{180}$ by 5-fold cross-validation with {\shadowgnn}~\cite{Zeng_etal/2021} and with three iterations of {\EEGL} (R1--R3) are given in Table~\ref{table:G180}. For {\EEGL} we present the results for vanilla and for random feature initialization (R0).
The parameter values were chosen with hyperparameter grid search.
{\EEGL} achieved an F1-score of 100\% with both vanilla and random initialization, considerably outperforming {\shadowgnn}. Note also that the difference between vanilla and random GNN (R0) is nearly 70\%. We speculate that this is due to the high 1-WL symmetry of $G_{180}$ and the relationship between the 1-WL and the target classes. 

In contrast to the graphs in Section~\ref{sec:old_motifs}, {\EEGL} needed more than one iteration to achieve an F1-score of 100\% on $G_{180}$. 
A closer look at the confusion matrices indicates that the difficulty is that each of the seven 1-WL equivalence classes is the union of four target classes.  
We illustrate the \textit{training dynamics} of (vanilla) {\EEGL} on one of the folds. 
In particular, for each iteration of {\EEGL} we analyse the confusion matrix summarizing {\EEGL}'s predictions for \textit{all} 180 nodes (i.e., for the training examples as well) and present some extracted top frequent patterns to demonstrate how {\EEGL} goes far beyond the node distinguishing power of 1-WL and hence, that of (vanilla) GNN.

Fig.~\ref{fig:G180_r0} shows the result obtained with vanilla GNN (R0). The four colors indicate the four predicted classes. 
Comparing the predictions with the true class values in Fig.~\ref{fig:G180_r3}, one can see that vanilla GNN achieves poor predictive performance. Somewhat surprisingly, it is even worse than that of 1-WL: While the nodes with distance 0, 2, and 4 from $C_{12}$ are 1-WL distinguishable, vanilla GNN cannot distinguish among them. 
We note that the target classes are colored consistently in all graphs in Figs.~\ref{fig:G180_r0}--\ref{fig:G180_r3}.
There is a similar problem with the nodes with distance 1 and 5 from $C_{12}$.
Furthermore, all nodes e.g. of distance 6 from $C_{12}$ are misclassified in R0. 
All top frequent patterns extracted from the local explanations in R0 are star graphs, with roots formed by their center nodes. 
Since patterns are matched with rooted subgraph isomorphism, these patterns define lower bound constraints on the nodes' degree.   
Still, already these ``weak'' patterns suffice to improve the predictive performance in the first iteration R1 (see Fig.~\ref{fig:G180_r1}). In fact, the class predictions obtained in R1 coincide with the 1-WL node partitioning. 
The top frequent patterns extracted from the local explanations in this round were semantically more meaningful, compared to those in R0. 
As an example, consider the rooted pattern $(P_1, r_1)$ in Fig.~\ref{fig:G180_motif_1}. 
While for any node $v$ of $G_{180}$ with distance 0 or 6 from $C_{12}$, there is a rooted subgraph isomorphism from $(P_1,r_1)$ to $(G_{128}, v)$, such a rooted subgraph isomorphism does not exist e.g. for the nodes with distance 5 from $C_{12}$.
On the other hand, from $(P_2,r_2)$ in Fig.~\ref{fig:G180_motif_2} there is a (resp. there is no) rooted subgraph isomorphism to $(G_{128}, v)$ for all nodes $v$ with distance 0 (resp. 6) from $C_{12}$.
Thus, the combination of these two patterns can distinguish the nodes on $C_{12}$ from those with distance 6 from $C_{12}$. 
Using the top frequent patterns extracted in R1 for feature annotation, a much better prediction is obtained in R2 (see Fig.~\ref{fig:G180_r2}). 
In particular, all nodes belonging to a copy of $M_0$ (Fig.~\ref{fig:AA}) or $M_3$ (Fig.~\ref{fig:BB}) are correctly classified. 
Note that these two motifs are non-isomorphic to each other as well as to $M_1$ and $M_2$. The nodes of $M_1$ (Fig.~\ref{fig:AB}) and $M_2$ (Fig.~\ref{fig:BA}) are, however, incorrectly classified. 
Though they are isomorphic motifs, there is no isomorphism between them mapping the bottom nodes (i.e., which lie on $C_{12}$) to each other.    
For the four occurrences of $M_1$ and for those of $M_2$ it holds that their nodes with the same distance from $C_{12}$ are predicted with the same target class. Recall that they are 1-WL indistinguishable. Still, a combination of the top frequent patterns extracted in R2 is able to overcome this problem in the next iteration R3, obtaining the correct classification (Fig.~\ref{fig:G180_r3}).
As an example, one can check that the nodes of $M_1$ and $M_2$ with distance 6 from $C_{12}$ in $G_{128}$ become distinguishable  by the rooted pattern in Fig.~\ref{fig:G180_motif_3}.
The example shows that it is possible for the predicted class to be incorrect for \textit{all} nodes of a block (see, e.g., the nodes on $C_{12}$ in Fig.~\ref{fig:G180_r0}).
Furthermore, while a block can remain unchanged after an iteration, its associated class may change (see, e.g., the nodes with distance 6 from $C_{12}$ in Figs.~\ref{fig:G180_r0} and \ref{fig:G180_r1}). In all such cases in the example considered, the new predicted class improves the accuracy. 

\textbf{Answers to Q1--Q3} In summary, for graphs that have high 1-WL symmetry and their 1-WL classes are formed by unions of target classes, the following are answers to the questions in Sect.~\ref{sec:introduction}.
Regarding Q1, \textit{{\EEGL} is capable of iterative self-improvement from explanations}, but, unlike the graphs in Sect.~\ref{sec:old_motifs}, it needs more iterations.
For Q2,  \textit{feature initializations may affect the number of iterations}. Running {\EEGL} with random initialization required less iterations than vanilla.
Also, {\EEGL} \textit{considerably outperformed} {\shadowgnn} in predictive performance.
Finally, regarding Q3, it can handle this kind of relationships between 1-WL classes and (target) label partitions as well, implying that \textit{{\EEGL} has a distinguishing power that goes beyond the limitations of vanilla GNNs}.

\subsection{Fullerenes}
\label{sec:fullerenes}

\begin{figure*}[t]
	\centering
	\begin{subfigure}[b]{0.2\textwidth}
		\centering
		\includegraphics[width=1\textwidth]{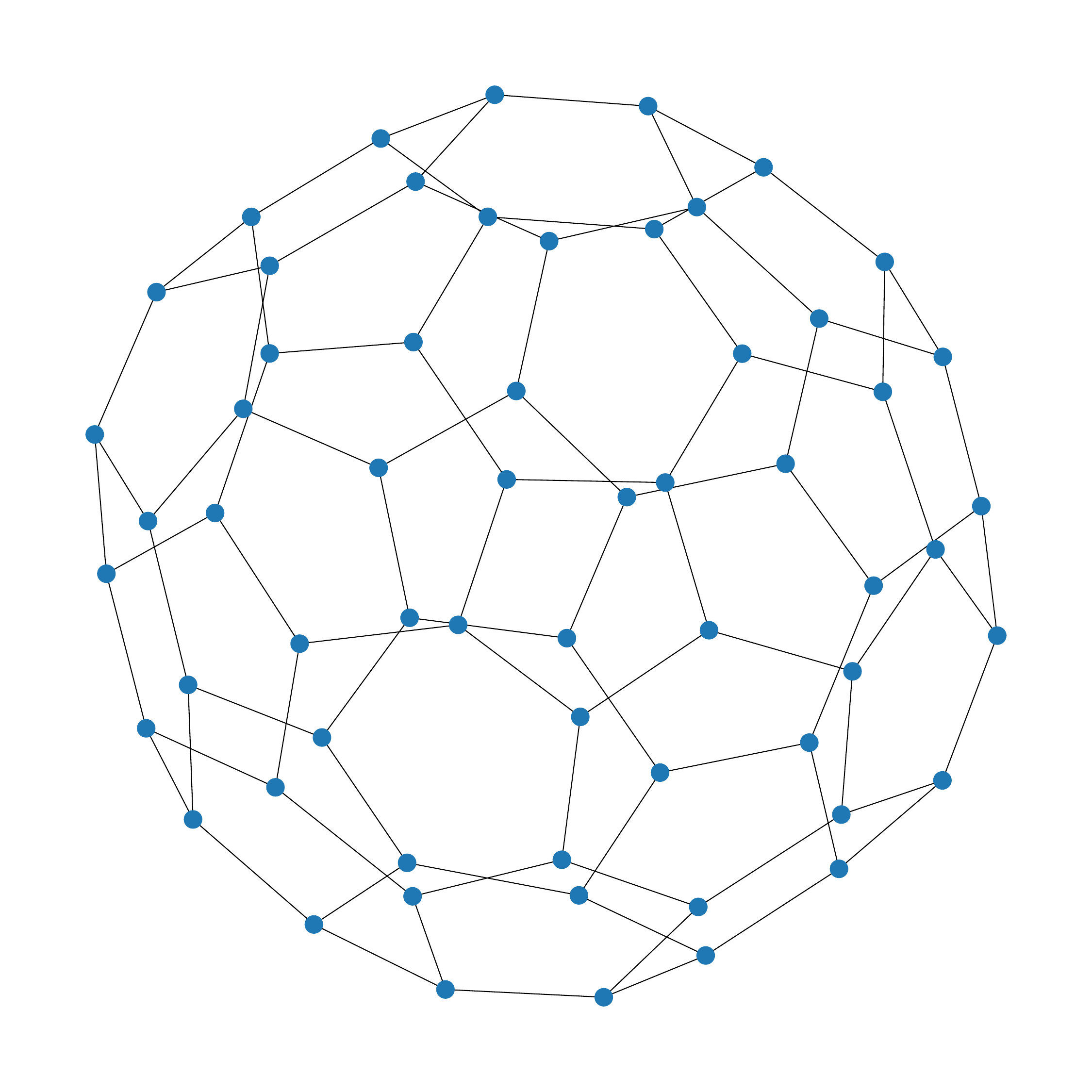}
		\caption{\centering $\text{C}_{60}$}
		\label{fig:C60}
	\end{subfigure} \hfill
	\begin{subfigure}[b]{0.2\textwidth}
		\centering
		\includegraphics[width=1\textwidth]{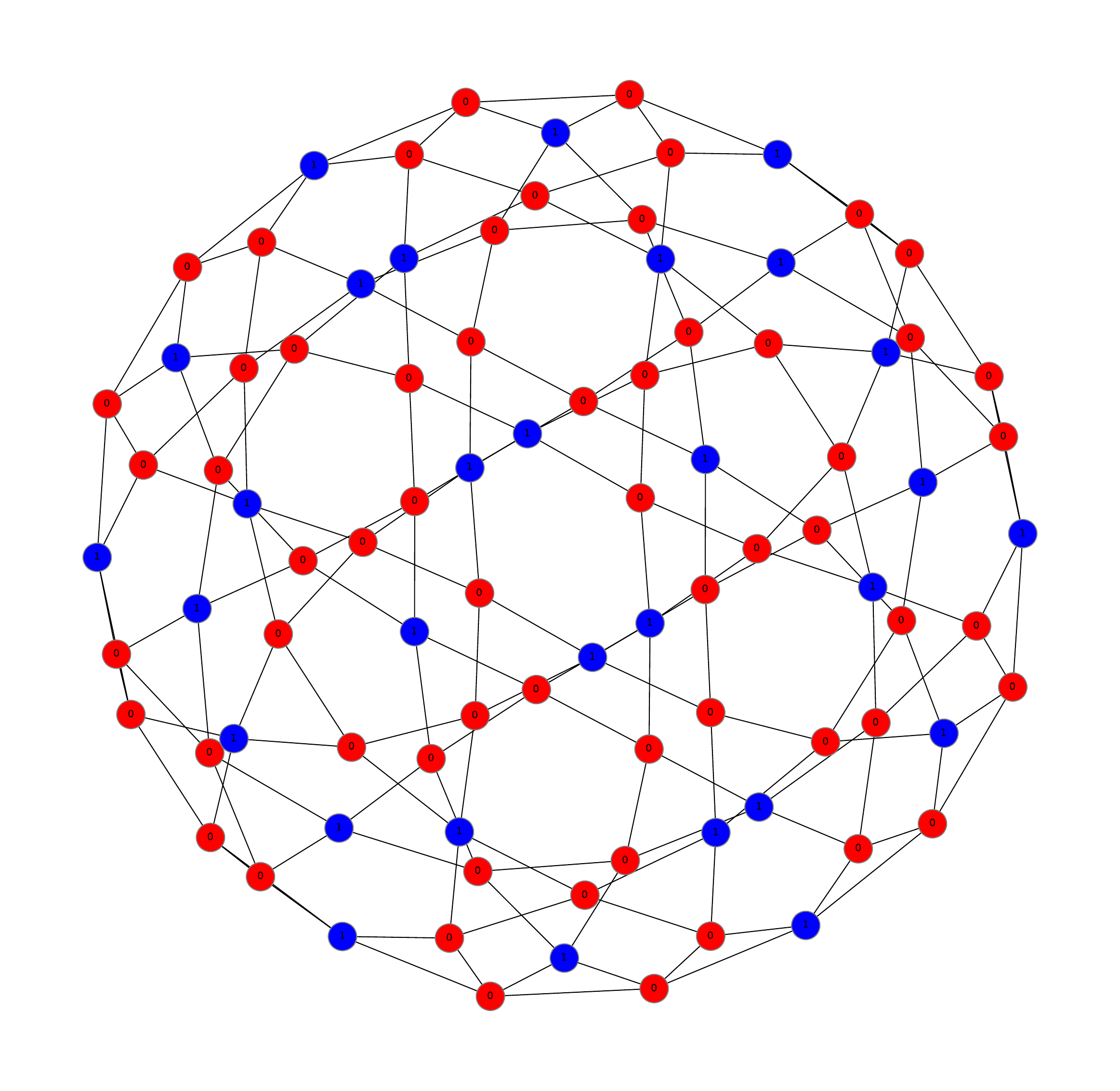}
		\caption{\centering linegraph of $\text{C}_{60}$}
		\label{fig:C60linegraph}
	\end{subfigure} \hfill
	\begin{subfigure}[b]{0.2\textwidth}
		\centering
		\includegraphics[width=1\textwidth]{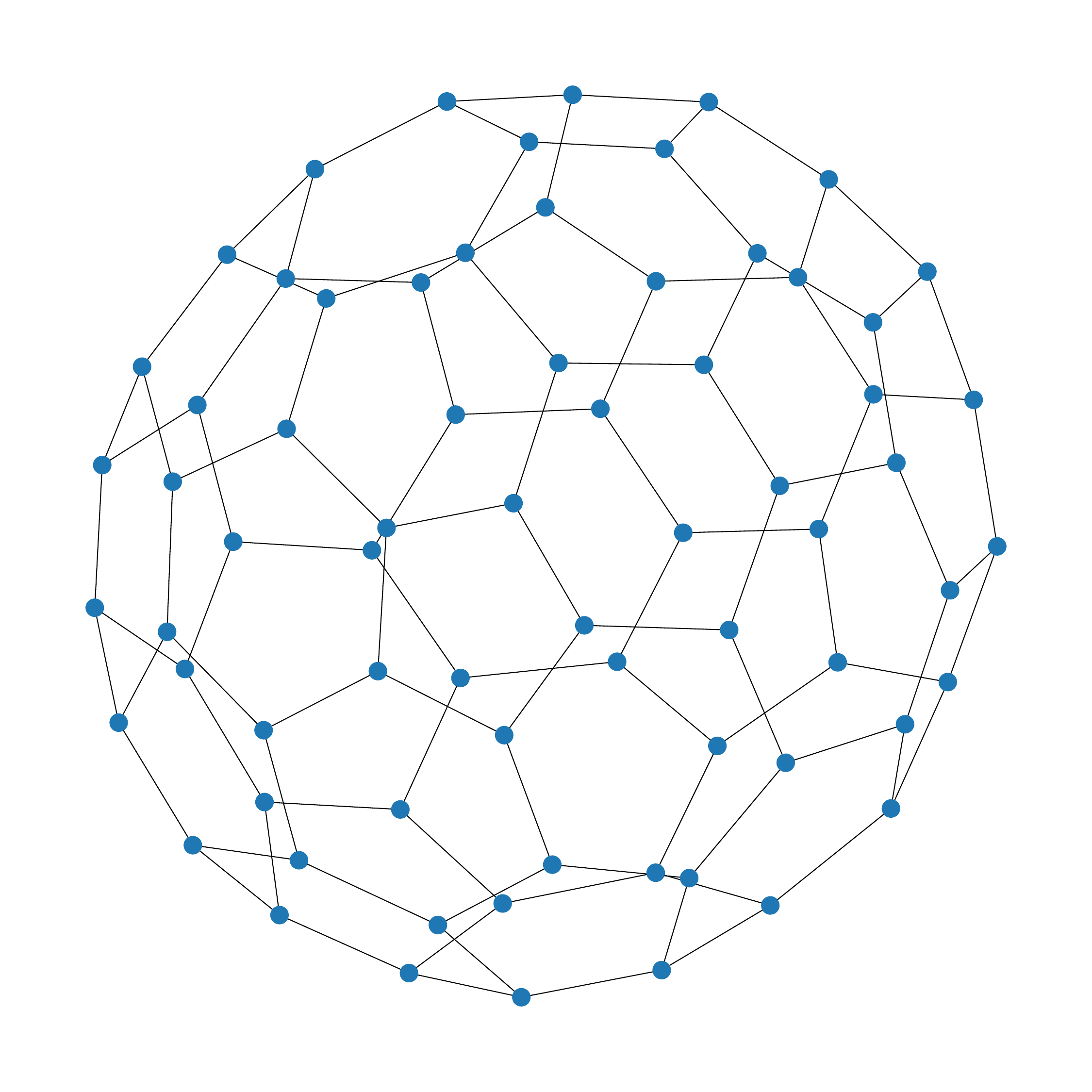}
		\caption{\centering  $\text{C}_{70}$}
		\label{fig:C70}
	\end{subfigure} \hfill
	\begin{subfigure}[b]{0.2\textwidth}
		\centering
		\includegraphics[width=1\textwidth]{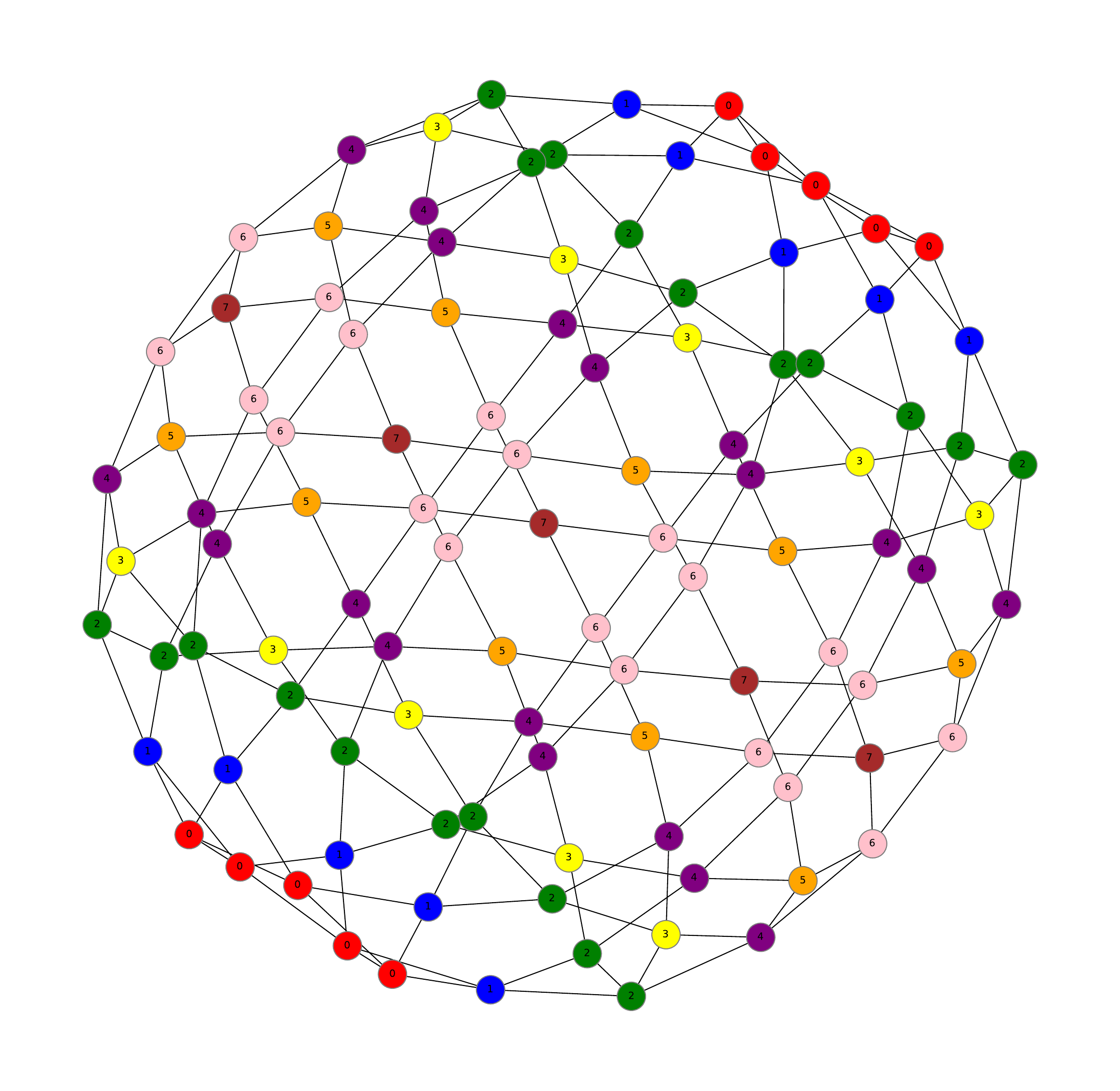}
		\caption{\centering linegraph of $\text{C}_{70}$}
		\label{fig:C70linegraph}
	\end{subfigure} \hfill
	
	\begin{subfigure}[c]{0.1\textwidth}
		\centering
		\includegraphics[width=1\textwidth]{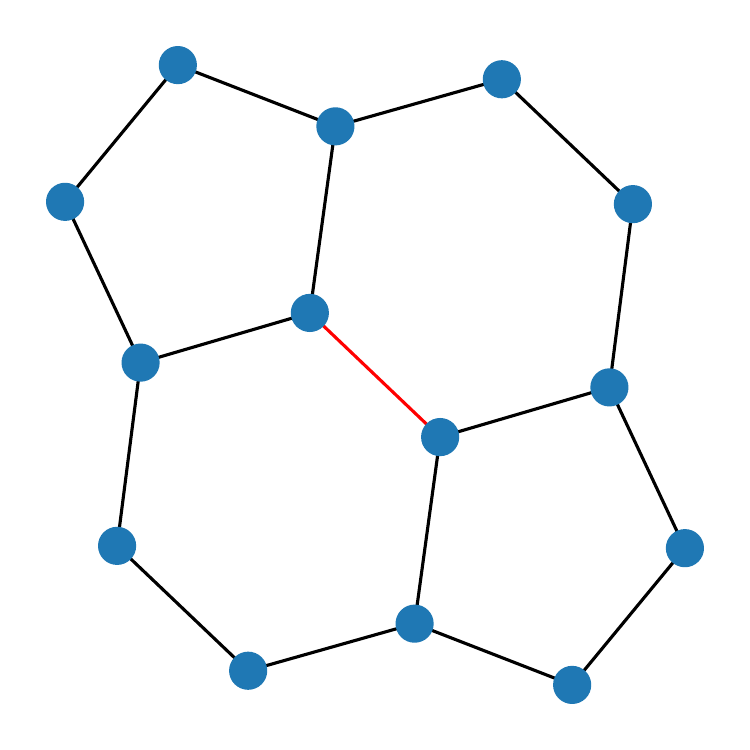}
		\caption{\centering T1}
		\label{fig:T1}
	\end{subfigure} \hfill
	\begin{subfigure}[c]{0.1\textwidth}
		\centering
		\includegraphics[width=1\textwidth]{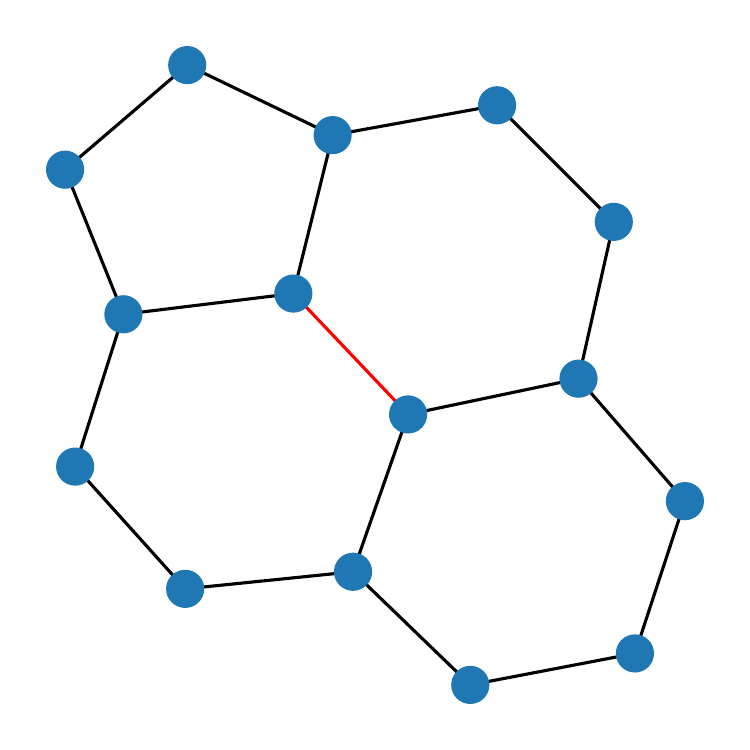}
		\caption{\centering T2}
		\label{fig:T2}
	\end{subfigure} \hfill
	\begin{subfigure}[c]{0.1\textwidth}
		\includegraphics[width=1\textwidth]{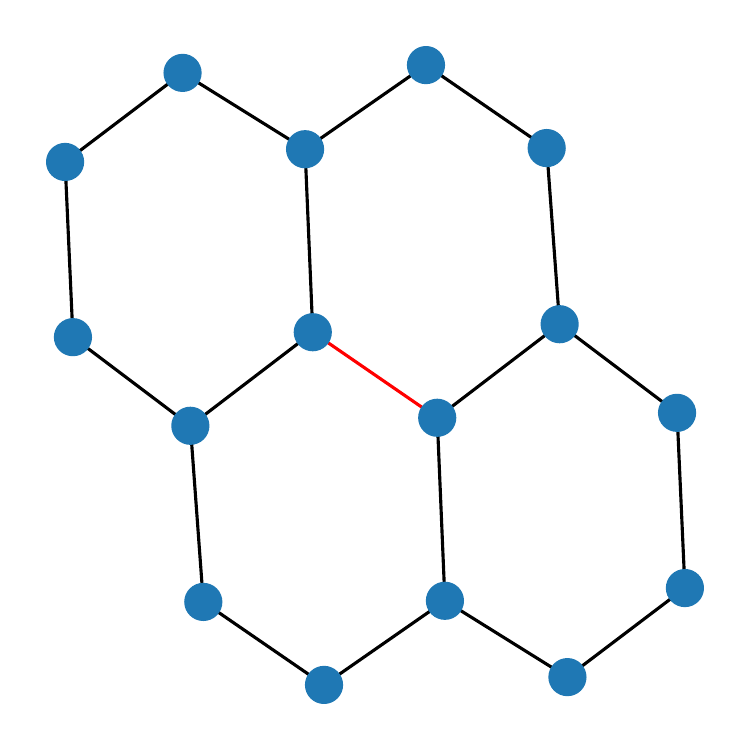}
		\caption{\centering T3}
		\label{fig:T3}%
	\end{subfigure} \hfill
	\begin{subfigure}[c]{0.1\textwidth}
	\includegraphics[width=1\textwidth]{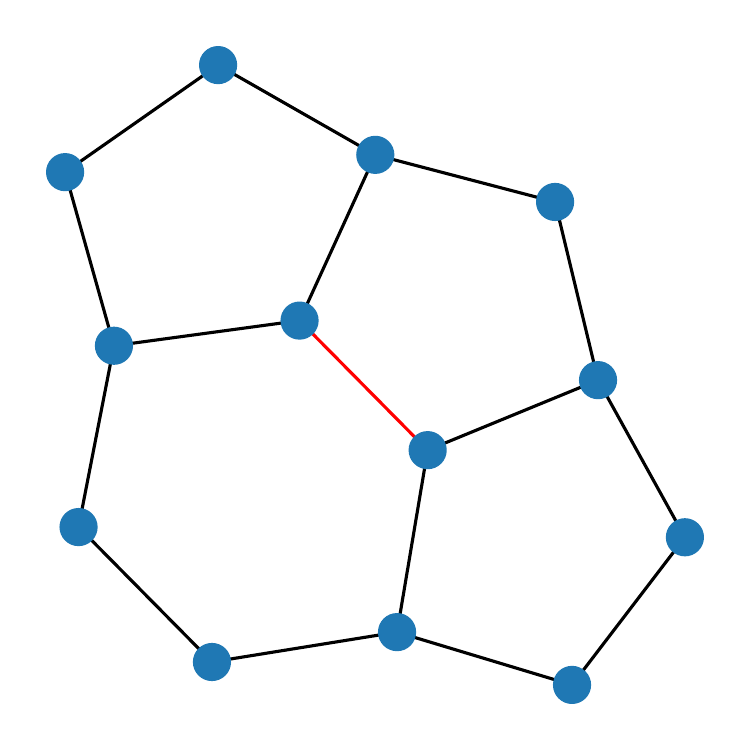}
	\caption{\centering T4}
	\label{fig:T4}%
    \end{subfigure} \hfill
	\begin{subfigure}[c]{0.1\textwidth}
	\includegraphics[width=1\textwidth]{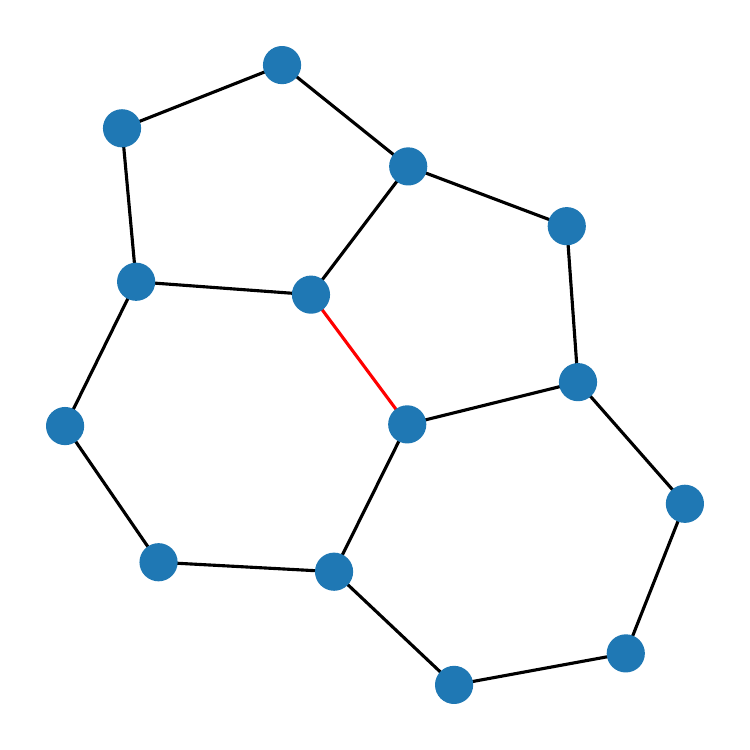}
	\caption{\centering T5}
	\label{fig:T5}%
    \end{subfigure} \hfill	
    \begin{subfigure}[c]{0.1\textwidth}
	\includegraphics[width=1\textwidth]{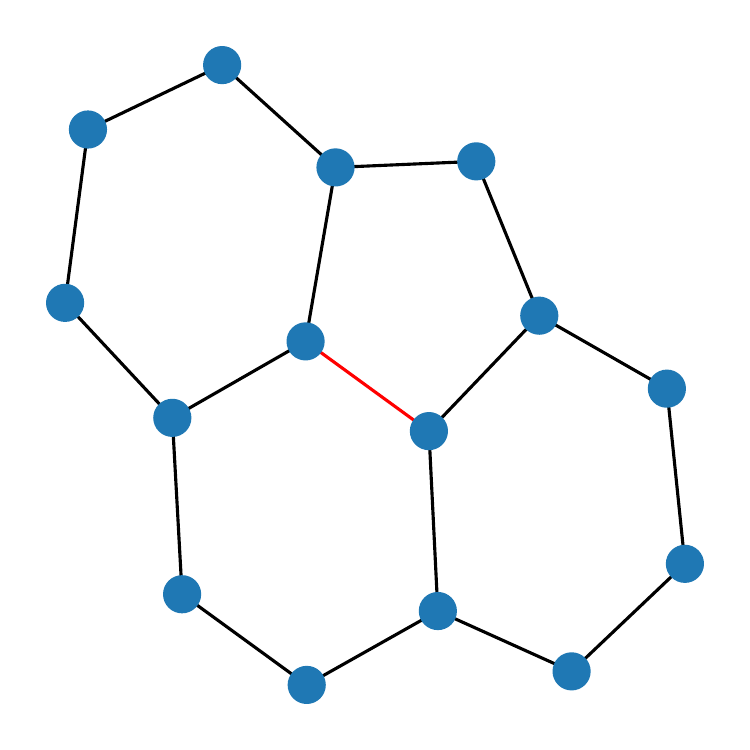}
	\caption{\centering T6}
	\label{fig:T6}%
    \end{subfigure} \hfill	
    \begin{subfigure}[c]{0.1\textwidth}
	\includegraphics[width=1\textwidth]{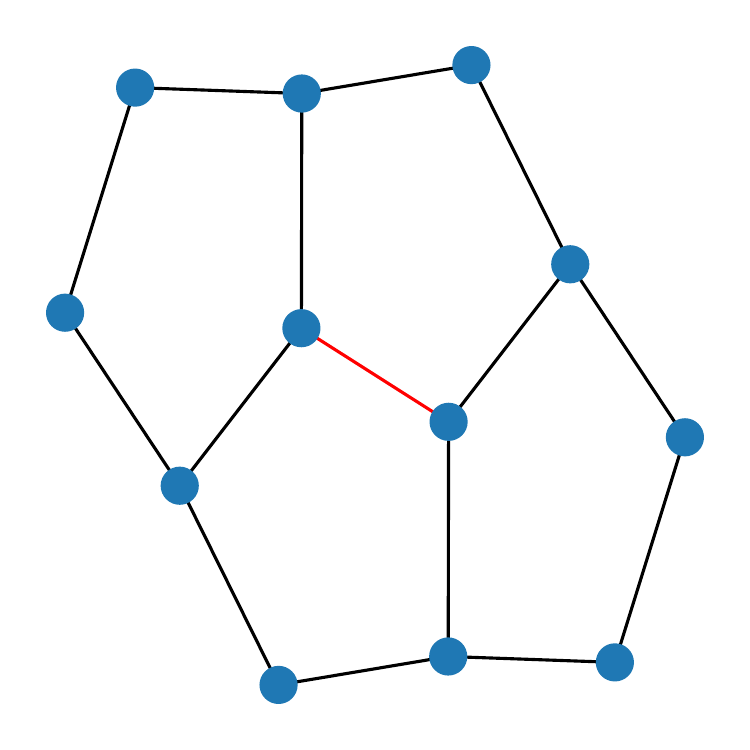}
	\caption{\centering T7}
	\label{fig:T7}%
    \end{subfigure} \hfill
    \begin{subfigure}[c]{0.1\textwidth}
	\includegraphics[width=1\textwidth]{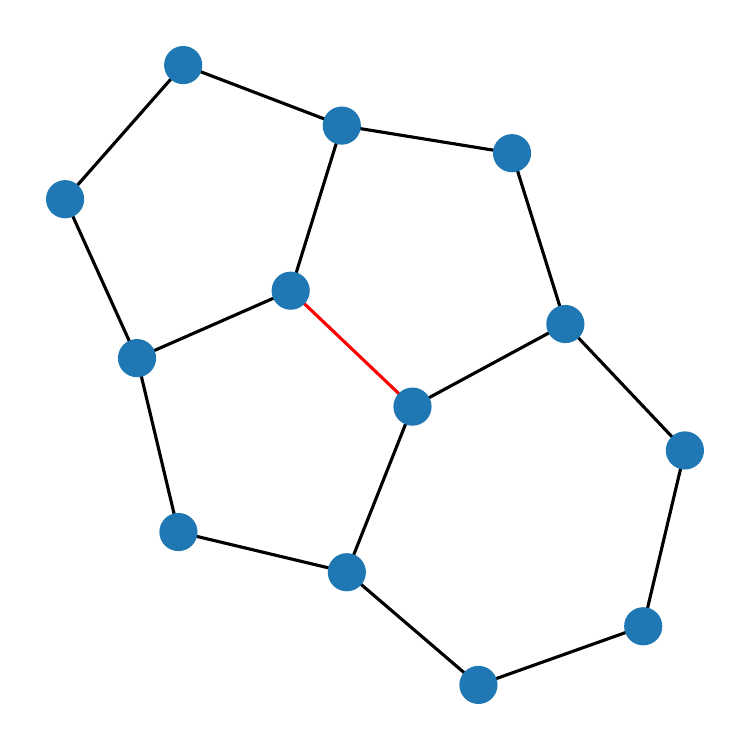}
	\caption{\centering T8}
	\label{fig:T8}%
    \end{subfigure} \hfill
    \begin{subfigure}[c]{0.1\textwidth}
	\includegraphics[width=1\textwidth]{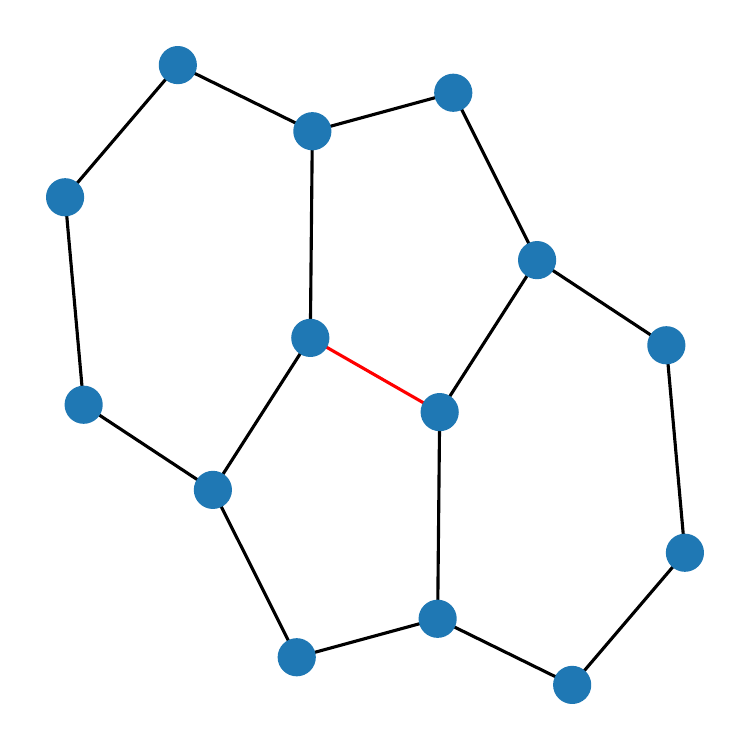}
	\caption{\centering T9}
	\label{fig:T9}%
\end{subfigure} \hfill\caption{Fullerenes $\text{C}_{60}$ (a) and its linegraph (b), $\text{C}_{70}$ (c), and its line graph (d). The colors in (b) and (d) indicate the corresponding bond types in (a) and (c), respectively. The different [6,6], [5,6], and [5,5] types of the bonds (in red) are given in (e)--(m).  
}
	\label{fig:fullerenes}
\vspace{-0.5em}
\end{figure*}

Now we turn to \textit{fullerenes}, a large class of 
molecules made of carbon atoms only,
having an atom-bond (graph) structure consisting of pentagons and hexagons~\cite{xing2022-fullerene}.
Examples are $\CA$ (the Buckminsterfullerene or ``buckyball'') and $\CB$ (see Figs.~\ref{fig:C60} and \ref{fig:C70}).
For every even  $n \geq 20$ except 22, fullerenes with $n$ carbon atoms contain 12 pentagonal and $n/2 -10$ hexagonal faces (e.g., $\CA$ is made of 12 pentagons and 20 hexagons). 
Their molecular graphs are 3-regular.


We consider bond (edge) classification problems. 
Target classes
are defined either by bond lengths or by 9 patterns (red edges in Figs.~\ref{fig:T1}--\ref{fig:T8} surrounded by different combinations of four polygons).
Edge classification is reduced to node classification by considering the line graphs of the atom-bond structure. The \textit{line graph} of a graph $G$ is a graph $G'$ such that each vertex of $G'$ represents an edge of $G$ and two nodes of $G'$ are connected by an edge iff their corresponding edges in $G$ share a common node (line graphs of $\CA$ and $\CB$ are in Figs.~\ref{fig:C60linegraph} and \ref{fig:C70linegraph}).
The line graphs of fullerenes are 4-regular.
Thus fullerene node and edge classification problems are of type (iii) in Section~\ref{sec:introduction} and
vanilla 1-WL cannot be used. Therefore in this section we use EEGL with \emph{random initialization}
(see Section~\ref{sec:old_motifs}). 


\begin{table}[t]
\begin{footnotesize}
\begin{tabular}{llllll}
\toprule  
fullerene & {\shadowgnn} & R0 & R1 & R2\\
\midrule
C24-D6d & 68.68$\pm$20.83& 84.29$\pm$24.91& 100.00$\pm$0.00& 100.00$\pm$0.00\\
C38-C1-3 & 36.70$\pm$9.77 & 36.81$\pm$9.08& 97.40$\pm$5.19& 97.45$\pm$5.09\\
C50-D5h-1 & 63.30$\pm$5.73& 72.46$\pm$18.14& 100.00$\pm$0.00& 100.00$\pm$0.00\\
C60-Ih & 57.08$\pm$4.76 & 92.06$\pm$4.54& 100.00$\pm$0.00& 100.00$\pm$0.00\\
C70-D5h & 56.87$\pm$8.27 & 62.99$\pm$11.18& 93.27$\pm$13.46& 100.00$\pm$0.00\\
C72-D6d & 59.24$\pm$5.33& 71.97$\pm$18.57& 100.00$\pm$0.00& 100.00$\pm$0.00\\
C74-D3h & 52.71$\pm$10.66& 64.24$\pm$18.03& 87.70$\pm$24.60& 87.70$\pm$24.60\\
C76-D2 & 63.50$\pm$8.41 & 66.00$\pm$7.94& 100.00$\pm$0.00& 100.00$\pm$0.00\\
C78-D3-1 & 51.82$\pm$10.54 & 59.78$\pm$10.90& 100.00$\pm$0.00& 100.00$\pm$0.00\\
C80-D5d-1 & 56.35$\pm$8.79 & 78.01$\pm$9.99& 100.00$\pm$0.00& 100.00$\pm$0.00\\
C82-C2-1 & 57.02$\pm$10.31 & 66.67$\pm$4.08& 100.00$\pm$0.00& 100.00$\pm$0.00\\
C84-D2-1 & 61.80$\pm$11.90& 72.19$\pm$15.93& 86.23$\pm$27.55& 86.23$\pm$27.55\\
C86-C1-1 & 55.29$\pm$7.50 & 53.30$\pm$13.40& 100.00$\pm$0.00& 100.00$\pm$0.00\\
C90-D5h-1 & 62.92$\pm$5.59 & 73.46$\pm$10.82& 100.00$\pm$0.00& 100.00$\pm$0.00\\
C94-C2-1 & 55.93$\pm$5.00 & 64.44$\pm$9.81& 100.00$\pm$0.00& 100.00$\pm$0.00\\
C96-D2-1 & 63.17$\pm$6.28 & 78.36$\pm$9.43& 100.00$\pm$0.00& 100.00$\pm$0.00\\
C98-C2-1 & 61.65$\pm$6.53 & 65.79$\pm$13.57& 100.00$\pm$0.00& 100.00$\pm$0.00\\
\bottomrule
\end{tabular}
\caption{
Average weighted F1-score results with SD in percentage obtained for different fullerenes with 5-fold CV with {\shadowgnn}~\cite{Zeng_etal/2021} and random initialization (R0), and for two iterations of {\EEGL} (R1, R2).
}
\label{table:fullerenes}
\end{footnotesize}
\vspace{-1cm}
\end{table}
\textbf{Experimental Results and Analysis} The results 
for different fullerenes\footnote{https://nanotube.msu.edu/fullerene/fullerene-isomers.html} are given in Table~\ref{table:fullerenes}. 
The results show that {\EEGL} (R2) has a high predictive performance on fullerenes and significantly outperforms {\shadowgnn}~\cite{Zeng_etal/2021}. 
Table~\ref{table:fullerenes} includes also $\textrm{C}_{60}$  and $\textrm{C}_{70}$ (see Figs.~\ref{fig:C60} and \ref{fig:C70}).
Of the 9 patterns in Fig.~\ref{fig:fullerenes} only two (T1 and T6) occur in $\textrm{C}_{60}$.
Edge partitions by these patterns coincides with that by the two bond lengths in $\textrm{C}_{60}$ (cf. Fig.~\ref{fig:C60linegraph}).
{\EEGL} achieved perfect classification in one iteration (R1) on $\textrm{C}_{60}$. Interestingly, instead of T1 and T6, {\EEGL} extracted two semantically correct, more compact rules distinguishing these two classes: An edge is incident to the node of a pentagon (cf. T1), resp., an edge lies on a pentagon (cf. T6).
On $\CB$, {\EEGL} achieved 100\% in R2.

For $\textrm{C}_{70}$, we also considered classifying edges by their length. While the number of target classes defined by the 9 structural patterns is only 4, the edges are partitioned into 8 blocks according to their bond lengths (Fig.~\ref{fig:C70linegraph}).  
{\EEGL} achieved an F1-score of 100\% in one iteration.
It is interesting that the patterns extracted for this problem had a much weaker distinguishing power than those obtained for $\textrm{C}_{60}$, still, this incomplete information was sufficient for GNN to learn a perfect model. This is an example of the possibility, mentioned in Section~\ref{sec:introduction}, of making progress based on weak explanation.

\textbf{Answers to Q1--Q2} In summary, \textit{{\EEGL} is capable of self-improve\-ment from explanations} in one iteration (Q1) and \textit{significantly outperforms} {\shadowgnn} on this kind of learning task as well (Q2). 


\section{Concluding Remarks}
\label{sec:conclusion}

We introduced EEGL, an iterative XAI-based model improvement approach to extend MPNN using frequent subgraph mining of explanation subgraphs to obtain features for improving predictive performance. 
The approach produced encouraging initial results, and it poses many directions for further research.



As we focused on graph-theoretic structural aspects, we have not considered node feature explanations.
In future work we will consider this extension.
A general question is to consider explanations from restricted tractable classes. This can be achieved either in the explainer (by modifying the denoising procedure in the case of GNNExplainer), or in the frequent subgraph miner. 
Another general question, continuing the list 
from the introduction is the following.

Q4: Does improved prediction performance in the iterations imply improved explanation quality?

There are several metrics for evaluating GNN explanations \cite{Sourav23}.
The evaluation of explanations in XAI is a difficult issue, with results on the problematic nature of explanations produced (see, e.g. \citep{Adeb18}). The lack of ground truth exacerbates the difficulties. Considering synthetic or real-world problems where ground truth is available may be useful, but~\cite{Faber21} warns of possible pitfalls. Considering node feature explanations  
could also be useful in this context as well.





\bibliography{ms}
\bibliographystyle{abbrv}

\newpage
\appendix
\begin{center}
\bf
Supplementary Materials for the Submission Titled \\[1em]
Iterative Graph Neural Network Enhancement via Frequent Subgraph Mining of Explanations
\end{center}


In Appendix~\ref{sec:background} we collect the necessary background about graph neural networks, the {\GNNExplainer} system, and frequent connected subgraph mining. Appendix~\ref{sec:eegl_app} gives a high-level pictorial representation of the EEGL framework. In Appendix~\ref{sec:patternextractionmodule} we describe the \emph{pattern-extraction module} which is a pre-requisite step for the \emph{top-k} pattern filtering and the annotation phases. In Appendix~\ref{sec:detailedresults} we report the detailed experimental results obtained for $M_2'$ (Fig.~\ref{fig:motifs}e in the submission) for all 10 folds. After the review process is complete, the source code will be publicly shared on Github.

\section{Background}
\label{sec:background}
For necessary background on GNN, {\GNNExplainer}, the Weisfeiler-Leman algorithm, and frequent subgraph mining, the reader is referred to \cite{hamilton17-representation,nijssen05-gaston,ying19-gnn_explainer}.

\paragraph{Graph Neural Networks (GNN)} 
We give a simplified description of GNN, corresponding to the type of networks used in {\GNNExplainer}~\cite{ying19-gnn_explainer}. In this paper we consider node classification. 
A GNN model $\Phi$ has as its inputs a graph $G = (V, E)$ and a feature matrix 
$X \in \mathbb{R}^{d \times |V|}$, where $X_v \in \mathbb{R}^d$ is a 
$d$-dimensional feature vector associated for every node $v \in V$. It computes the representation $z_v$, the \emph{embedding} of each node $v \in V$, via $L$ layers of \emph{neural message passing} as follows. Starting with initial node features $h_i^{0} = X_{v_i}$ for every $v_i \in V$, repeat the following three steps for each layer $l \in [L]$:
\begin{itemize}
\item[(i)] Compute neural messages $m_{ij}^l = \text{MSG}(h_i^{l-1}, h_j^{l-1}, r_{ij})$ for every pair of nodes $(v_i, v_j) \in V$, where \text{MSG} is the function which computes the message passed from $v_j$ to $v_i$, and $r_{ij}$ is a relation between $v_i$ and $v_j$, e.g., the edge relation. Usually the message is $h_j^{l-1}$ if $(v_i, v_j) \in E$, else 0. \newline
\item[(ii)] For each node $v_i$ aggregate the messages from its neighborhood $\mathcal{N}_{v_i}$ as $M_i^l = \text{AGG}({m^l_{ij} |  v_j \in \mathcal{N}_{v_i}})$, where AGG is the function used to aggregate all the messages passed from the neighborhood. AGG should be invariant or equivariant to the permutations of its inputs. \newline
\item[(iii)] For every node $v_i$ update its features for using the aggregated messages from the neighborhood and the previous features as $h_i^l = \text{UPDATE}(M_i^l, h_i^{l-1})$, where UPDATE is the function used to combine the two inputs.
\end{itemize}
After the computation through the $L$ layers, the embedding for every node $v_i$ is the features computed for the $L$'th layer i.e. $z_{v_i} = h_i^L$.
For a fully supervised node classification task, the training loss is $\mathcal{L} = \sum_{v \in \mathcal{V}_{train}}^{}{-\text{log}(\text{softmax}(z_v, y_v))} $, where $y_v \in \{0,1\}^c$ is \emph{one-hot} vector indicating the ground truth label for node $v$, $c$ is number of labels, and $\mathcal{V}_{train}$ is the set of nodes in the training set.

\paragraph{\GNNExplainer}~\citep{ying19-gnn_explainer} was the first general model-agnostic approach specifically designed for post-hoc explanations on graph learning tasks. Given a graph G, and a query in the form of a node in the graph, it identifies a compact subgraph structure and a small subset of node features that show significant evidence in playing a role in the prediction of the query node. The GNNExplainer procedure accomplishes this by formulating the explanation generation problem as an optimization task that maximizes the mutual information between a GNN's prediction and 
subgraph structures.

\textbf{Frequent Subgraph Mining}
%
Frequent subgraph mining, a well-established subfield of data mining, 
is concerned with the following problem: \textit{Given} some finite (multi)set $\cD$ of graphs and a frequency threshold $\tau \in (0,1]$, \textit{generate} all \textit{connected} graphs that are subgraphs of at least $\lceil \tau |\cD|\rceil$ graphs in $\cD$.
In the {\EEGL} system 
we use {\Gaston}~\citep{nijssen05-gaston}, one of the most popular 
algorithms for the generation of frequent connected subgraphs. 
%
Similarly to most other heuristics designed for frequent subgraph mining, {\Gaston} is based on the generate-and-test paradigm. 
One if its distinguishing features compared to other frequent subgraph mining algorithms is that it enumerates frequent patterns in accordance with a structural hierarchy (path, trees, and cyclic graphs), where cyclic patterns are generated in a DFS manner, ``refining'' frequent patterns by extending them with some new edge.
Regarding its pattern matching component, i.e., which decides whether a candidate pattern is frequent or not, {\Gaston} utilizes that frequent patterns are generated in increasing size. 
Instead of testing subgraph isomorphism from scratch for the graphs in $\cD$, it stores the embeddings  of the frequent patterns already found and decides subgraph isomorphism for a candidate pattern by utilizing the embeddings calculated for its antecedents.

\section{shaDow-GNN}
\label{sec:shadowgnn}
Regular GCN's compute the node-features from a scope - which is a $k$-hop neighborhood of nodes - via a $k$-depth GCN. {\shadowgnn}~\cite{Zeng_etal/2021} argues that this leads to \emph{oversmoothing} of node-features, because mostly only a few nodes in the neighborhood of a node are relevant for the property/label of that node. This is especially worsened if the relevant nodes are somewhat farther in the graph, because then we need a large $k$ to get them into the scope. Hence the entire $k$-hop neighborhood becomes too big a scope for computing node features, and it becomes hard to discern useful information from the few relevant nodes among too much noise from the sea of irrelevant nodes. They argue that the scope should only be a subgraph of the $k$-hop neighborhood, and the depth $l$ of a GCN should be decoupled from the scope of a node . With this decoupling, {\shadowgnn} can have GCN's with any depth over the reduced scope which should fix the oversmoothing of the node features. However this introduces a new problem of identifying a good scope for the nodes. They propose methods using random-walk, personalised page rank(PPR) etc. to filter the scope from the neighborhood.

\section{Explanation Enhanced Graph Learning}
\label{sec:eegl_app}

\begin{figure*}
    \centering
    \includegraphics[width=0.8\textwidth]{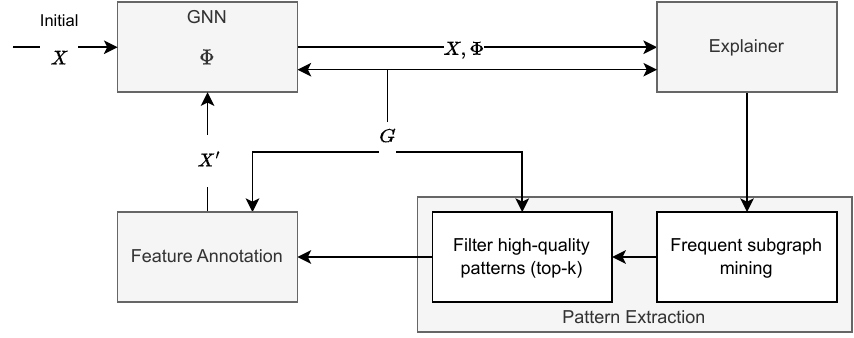}
    \caption{High-level depiction of the EEGL Process}
    \label{fig:eegl}
\end{figure*}

Figure~\ref{fig:eegl} shows a high level abstraction of the four major phases of the EEGL framework. We also note that the \emph{pattern extraction} module consists of two sub-modules
\begin{enumerate*}
    \item Frequent-subgraph mining
    \item Top-k pattern filtering
\end{enumerate*}

\section{Pattern Extraction Module}
\label{sec:patternextractionmodule}

For a class label $\classlabel \in \setofclasslabels$,  {\sc Maximal\_Frequent\_Pattern\_Mining} (line~\ref{line:maximal_frequent_patterns} of Alg.~\ref{alg:EEGL}) computes a set of maximal frequent rooted patterns in $\mathcal{E}_c$ in two steps:
(i) It generates a set of frequent rooted subgraphs of $\mathcal{E}_c$ and (ii) selects the maximal rooted patterns from this set.

Regarding (i), the following frequent pattern mining problem is solved: 
\textit{Given} $\mathcal{E}_c$ containing $m$ explanation graphs for some $m \geq 0$ integer and a frequency threshold $\tau \in (0,1]$, \textit{enumerate} the set of rooted patterns $(P,r)$ such that 
$(P,r)$ is frequent, i.e., there is a set $\mathcal{E}'_c \subseteq \mathcal{E}_c$ of rooted explanation graphs such that $|\mathcal{E}'_c| = \left\lceil \tau m\right\rceil$ and for all $(P',r') \in \mathcal{E}_c'$ there exists a rooted subgraph isomorphism from $(P,r)$ to $(P',r')$.
Note that $\mathcal{E}_c$ may contain explanation graphs that are not associated with a root. While such explanation graphs do not support any of the frequent rooted patterns, they have an impact on the rooted patterns' (relative) frequencies. 
The above problem is solved by {\Gaston}~\cite{Nijssen/Kok/04} as follows: For all rooted explanation graphs $(P,v) \in \mathcal{E}_c$, $v$ is associated with a distinguished node attribute label. Running {\Gaston} on these modified explanation graphs with frequency threshold $\tau$, it returns a set of frequent subgraphs. From this set we keep only the connected components of the frequent subgraphs that contain a node with the  distinguished attribute label. This node will be regarded as the root of the pattern.

Regarding (ii), we remove all rooted patterns returned in the previous step that have a rooted subgraph isomorphism to some other rooted pattern and return the remaining rooted pattern set.

\section{Results}
\label{sec:detailedresults}
In this section we present the detailed results obtained for $M_2'$ (Fig.~\ref{fig:motifs}e in the submission). For each of the 10 folds, we present the 
\begin{itemize}
    \item  the weighted F1-score results in percentage (top left table) obtained with the label encoding, random, and adversarial settings, as well as after the three iterations of {\EEGL} (Round-0, Round-1, Round-2),
    \item the frequencies of the node labels in the test data (top right table),
    \item the confusion matrices obtained with the label encoding, random, and adversarial settings, as well as after the three iterations of {\EEGL} (Round-0, Round-1, Round-2),
    \item the $d=10$ maximal frequent subgraphs extracted by {\EEGL} in the first (R0 $\to$ R1) and the second (R1 $\to$ R2) iteration. (Class labels are indicated on the top. Note that we can have more than one pattern for a node label.) 
\end{itemize}



\def\foldresultspath{24/01/16/appendix_p007-02}
\foreach \var in {01,02,03,04,05,06,07,08,09,10} {
  
  \newpage
  \subsection*{Fold \var}
  \begin{table}[!h]
      \input{figures/results/\foldresultspath/tab-\var}
      \caption{F1-Scores for fold-\var}
  \end{table}
  \begin{figure}[!h]
    \includegraphics[width=\linewidth]{figures/results/\foldresultspath/cm_fold-\var}
    \caption{Confusion Matrix for fold-\var}
  \end{figure}
  \begin{figure}[!h]
    \includegraphics[width=\linewidth]{figures/results/\foldresultspath/fsg_fold-\var}
    \caption{Top-K patterns for fold-\var from \emph{round-0} explanations. These patterns were used to annotate features for \emph{round-1}}
  \end{figure}

}

\newpage




\end{document}